\documentclass[10pt,journal,compsoc]{IEEEtran}

\ifCLASSOPTIONcompsoc
  \usepackage[nocompress]{cite}
\else
  \usepackage{cite}
\fi
\usepackage{bm}
\usepackage{mathrsfs}
\usepackage{textcomp}
\usepackage{epsfig}
\usepackage{amsthm}
\usepackage{dsfont}
\usepackage{indentfirst}
\usepackage{amsmath}
\usepackage{amssymb}
\usepackage{subfigure}
\usepackage{multirow}
\usepackage{graphicx}
\usepackage{fancyhdr}
\usepackage{epsfig}
\usepackage{mathrsfs}

\usepackage{cite}
\usepackage{color}
\usepackage{float}
\usepackage{algorithm}
\usepackage{algpseudocode}
\usepackage{amsmath}

\begin{document}

\title{Deep Convolutional Neural Network for Multi-modal Image Restoration and Fusion}

\author{Xin~Deng,~\IEEEmembership{Student member,~IEEE,}
        and~Pier Luigi~Dragotti,~\IEEEmembership{Fellow,~IEEE}
\IEEEcompsocitemizethanks{\IEEEcompsocthanksitem The authors are with the Department
of Electrical and Electronic Engineering, Imperial College London, London,
UK.\protect\\
E-mail: x.deng16, p.dragotti@imperial.ac.uk
}
\thanks{Xin Deng is supported by Imperial-CSC scholarship.}}

\markboth{Journal of \LaTeX\ Class Files}%
{Shell \MakeLowercase{\textit{et al.}}: Bare Demo of IEEEtran.cls for Computer Society Journals}

\IEEEtitleabstractindextext{%
\begin{abstract}
In this paper, we propose a novel deep convolutional neural network to solve the general multi-modal image restoration (MIR) and multi-modal image fusion (MIF) problems. Different from other methods based on deep learning, our network architecture is designed by drawing inspirations from a new proposed multi-modal convolutional sparse coding (MCSC) model. The key feature of the proposed network is that it can automatically split the common information shared among different modalities, from the unique information that belongs to each single modality, and  is therefore denoted with CU-Net, i.e., Common and Unique information splitting network. Specifically, the CU-Net is composed of three modules, i.e., the unique feature extraction module (UFEM), common feature preservation module (CFPM), and image reconstruction module (IRM). The architecture of each module is derived from the corresponding part in the MCSC model, which consists of several learned convolutional sparse coding (LCSC) blocks. Extensive numerical results verify the effectiveness of our method  on a variety of MIR and MIF tasks, including RGB guided depth image super-resolution,  flash guided non-flash image denoising, multi-focus  and multi-exposure image fusion. 
\end{abstract}

}

\maketitle

\IEEEdisplaynontitleabstractindextext

\IEEEpeerreviewmaketitle

\IEEEraisesectionheading{\section{Introduction}\label{sec:introduction}}
\IEEEPARstart{M}{ulti-modal} image processing has been attracting increasing interest from the computer vision community, due to a variety of intriguing  applications, e.g.,  image style transfer \cite{gatys2016image,johnson2016perceptual}, image fusion \cite{liu2016image,li2019densefuse}, RGB guided depth image super-resolution \cite{gu2017learning,liu2017robust}, image denoising  \cite{yan2013cross}. Based on the reconstruction target, these applications can be roughly classified into two categories, the multi-modal image restoration (MIR) and multi-modal image fusion (MIF)  task.
 Given an image $\bm{x}$ and another image $\bm{y}$, in MIR problem, one aims to recover a better version of $\bm{x}$ with the guidance of  $\bm{y}$, while the MIF task aims to fuse $\bm{x}$ and $\bm{y}$ to a new image $\bm{z}$ which has both the advantages of $\bm{x}$ and $\bm{y}$. Different from uni-modal image processing tasks, e.g., single image super-resolution,  multi-modal image processing usually requires proper modelling of the dependencies across different modalities.

Sparse coding and dictionary learning have been used widely to relate different modalities in multi-modal image processing.  In \cite{kwon2015data}, a method was proposed to learn one dictionary for each modality, and the dependencies across different modalities are modelled by assuming all the dictionaries share the same sparse representations.  In \cite{song2019multimodal},   a coupled dictionary learning algorithm was proposed to learn a pair of common and unique dictionaries for each modality, where only the common dictionaries share the same sparse representations. These methods based on sparse coding can provide explicit modellings, but the calculation of sparse codes is often time-consuming. Recent studies show that deep neural networks can also be used to correlate different modalities.  Li \textit{et. al} \cite{li2019joint} proposed a deep joint filtering network and Kim \textit{et. al} \cite{kim2018deformable} proposed a deformable kernel network. Both approaches use a two-stream convolutional neural network (CNN) architecture to extract the features from  two different modalities and then combine them through another CNN to achieve the final reconstruction. However, the dependencies modelled by them are quite implicit.  
\begin{figure*}
	\centering
	\centerline{\epsfig{figure=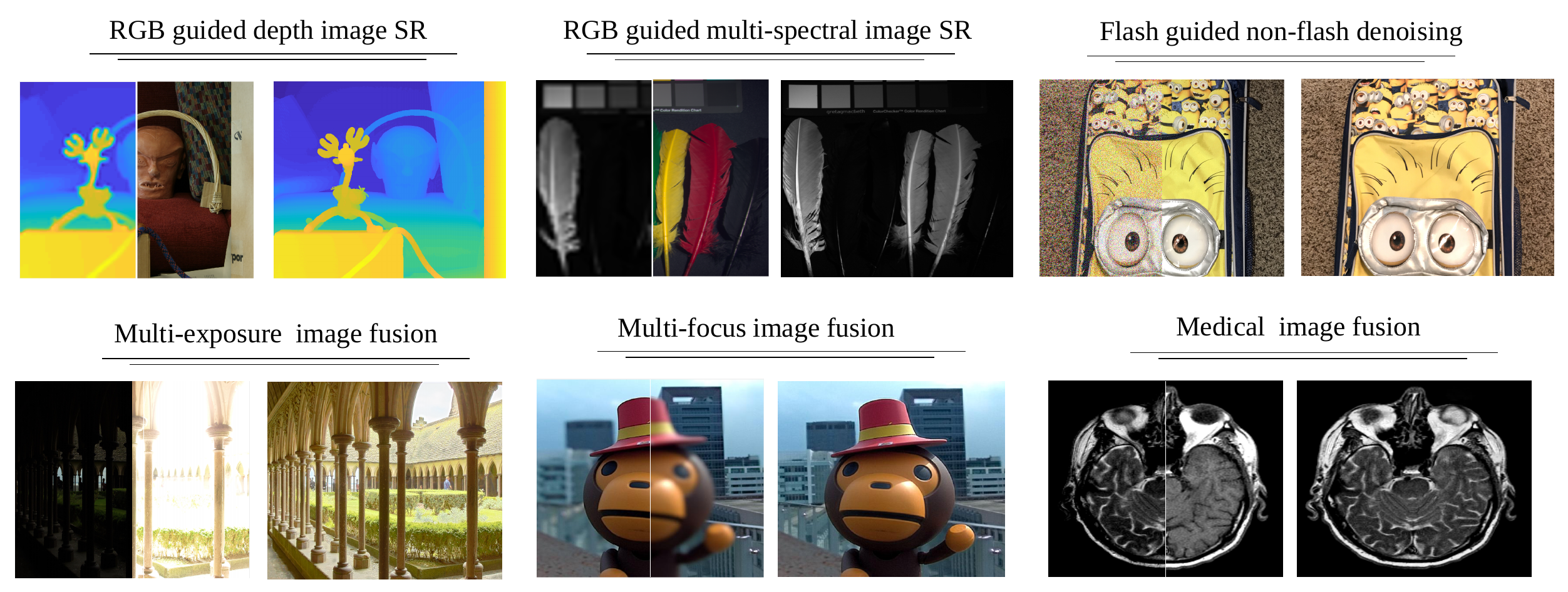, width=17cm}}
	\vspace{-1em}
	\caption{Examples of different multi-modal image restoration and fusion tasks. The first row shows the MIR  related applications, including RGB guided depth image SR, RGB guided multi-spectral image SR, and flash guided non-flash image denoising.  The second row shows the MIF related applications, including multi-exposure image fusion, multi-focus image fusion, and medical image fusion.  }\label{first}
	\vspace{-1em}
\end{figure*}
To address the aforementioned issues, the intuitive solution is to incorporate the sparse coding modelling into the neural network. Recently,  in \cite{deng2019deep}, the authors proposed a deep neural network which unfolded the iterative shrinkage and thresholding
algorithm (ISTA)  into a two-branch deep neural network, in order to solve the multi-modal image super-resolution problem. However, since the traditional sparse coding is performed at patch level, it may not be as effective as a CNN which exploits both neighbourhood information as well as global features in the image. Very recently, the authors in \cite{marivani2019learned} extent the traditional patch-based sparse coding to the convolutional sparse coding, and then unfold it into a deep convolutional neural network. However, it only targets the  multi-modal image super-resolution problem.


In this paper, we  aim  to solve the general multi-modal image restoration and fusion problems, by proposing a deep convolutional neural network named the Common and Unique information splitting network (CU-Net). To the best of our knowledge, this is the first time a universal framework is proposed to solve both the MIR and MIF problems. Compared with  other empirically designed networks, the proposed CU-Net  is derived from a new multi-modal convolutional sparse coding (MCSC) model, and thus each part of the network has good interpretability. 
The MCSC model is developed upon a recent coupled dictionary learning algorithm in \cite{song2019multimodal}. This algorithm is demonstrated to model well the dependencies across modalities, but it is performed at patch level, which means it only models the patch dependencies.  In contrast, our MCSC model is defined at image level, and thus we can model the image dependencies across modalities.
The main contributions of this paper can be summarized as follows:
\begin{itemize}
	\item  We propose a novel multi-modal convolutional sparse coding (MCSC) model, which has two variations to solve the general multi-modal image restoration (MIR) and multi-modal image fusion (MIF) problems, respectively. In this model, we represent each modality with two convolutional dictionaries: one for the common feature representation and the other for the unique feature representation. 
	
	\item  Based on the MCSC model, we design a deep convolutional neural network, i.e., CU-Net, to solve the MIR and MIF problems. Our network is able to automatically split the common features from the unique features between the target and guided modality,  which  is beneficial to  both MIR and MIF tasks.
	
	\item  We test the performance of the proposed method on various MIR and MIF tasks, such as RGB guided depth image super-resolution, flash guided non-flash image denoising,   multi-focus and multi-exposure image fusion.  The numerical results validate the effectiveness, flexibility, and universality of our method.
\end{itemize} 

The remainder of this paper is organized as follows. In Section \ref{rela}, we review the related work on MIR and MIF. In Section \ref{sparse}, we introduce two variations of the  multi-modal convolutional sparse coding model to solve the MIR and MIF problems, respectively. Based on this model, the proposed  Common and Unique information splitting network (CU-Net) is introduced  in Section \ref{dcus}. Finally, Section \ref{expe} shows the numerical results and Section \ref{con} concludes this paper.

\begin{figure*}
	\centering
	\centerline{\epsfig{figure=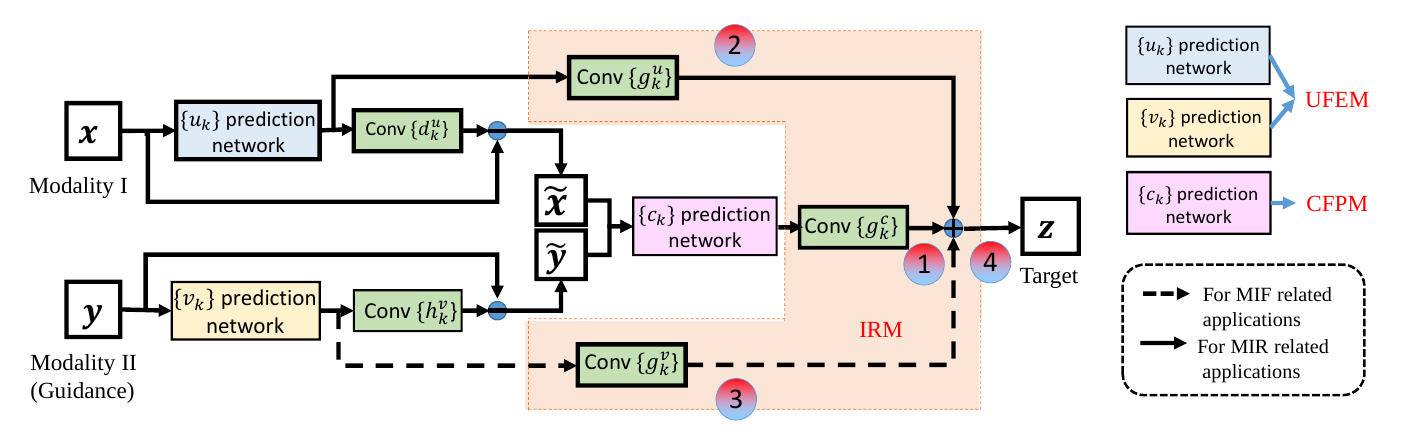, width=16cm}}
	\caption{Network Architecture of the proposed CU-Net. For the MIR related tasks, the final reconstruction (Point 4) is composed of the common reconstruction (Point 1) and the unique reconstruction  (Point 2). For the MIF related tasks, the final reconstruction is composed of the common reconstruction (Point 1) and the two unique reconstructions  (Point 2 and Point 3). }\label{frame}
\end{figure*}
\begin{figure}
	\centering
	\centerline{\epsfig{figure=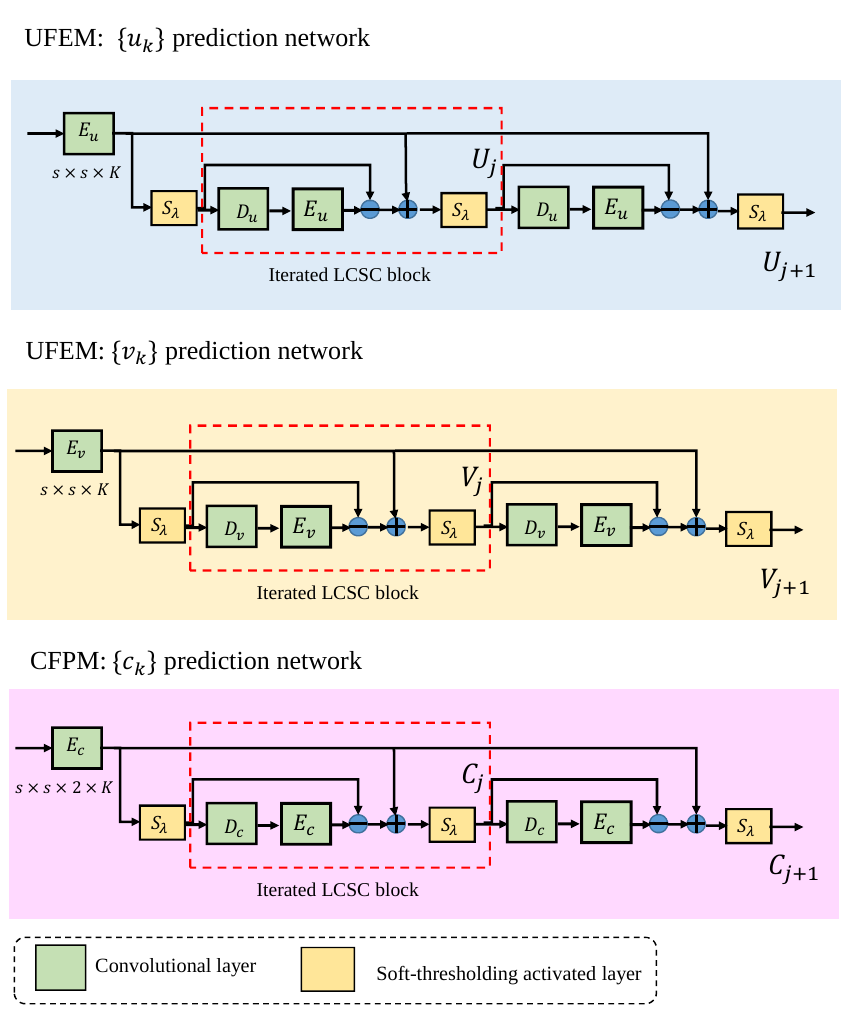, width=9cm}}
	\vspace{-1em}
	\caption{The network architectures of UFEM and CFPM.   }\label{module}
	\vspace{-2em}
\end{figure}

\section{Related Work} \label{rela}
\subsection{Multi-modal image restoration} 
For the MIR task, we first review  traditional methods based on guided image filtering, and then introduce more recent methods based on deep learning. The MIR methods based on guided image filtering aim to transfer the salient structures in the guidance image  to the target image, and they can be further classified into two categories: the statically guided \cite{kopf2007joint,he2010guided,ma2013constant,kou2015gradient, jevnisek2017co} and dynamically guided methods \cite{brox2008efficient,zhang2014rolling,shen2015mutual,shen2015multispectral,ham2018robust,guo2018mutually}.
The statically guided methods assume that the guidance image is not updated at each iteration, and thus they are only effective for the scenarios where the guidance image provides sufficient reliable details. Representative algorithms include the joint bilateral filter (JBF) \cite{kopf2007joint}, guided filter (GF) \cite{he2010guided}, and  weighted median filter (WMF) \cite{ma2013constant}. They may struggle when there are structural inconsistencies between the two images, e.g., RGB and depth images.   For the dynamically guided methods, they consider both the original guidance image and the filtered target image as guidance, which makes them more resilient to inconsistency between the two images. Specifically,  Zhang \textit{et al.} \cite{zhang2014rolling} proposed  a rolling guided filter (RGF), which iteratively used the filtered image as guidance for small structure and edge recovery. 
Ham \textit{et al.} \cite{ham2018robust} proposed a static/dynamic (SD) filter in which the static filter convolves the input image  with a weight function calculated from only the static guidance image, and the dynamic filter uses the weight function repeatedly calculated from the filtered output image. To avoid the transfer of inconsistent structures between guidance and target images, Shen \textit{et al.} \cite{shen2015mutual} defined three different types of structures: mutual structure, inconsistent structure and smooth regions. They further defined a normalized cross-correlation (NCC) metric to find the mutual structures between two images and used them to guide the filtering. However, the NCC metric is defined at patch level, and thus it can only  measure the patch similarity and sometimes leads to halo artefacts. To solve this problem, Guo \textit{et al.} \cite{guo2018mutually} recently proposed a mutually guided image filter (muGIF) which defines a relative structure to measure the similarity  between two images at image level. 

Recently, some works proposed to use deep neural networks to solve the MIR problem. The related methods can be classified into two categories: methods aiming to solve the general MIR problem and methods that are tailored to solve the MIR problem for specific pairs of modalities (e.g., RGB and depth). For the general MIR task, Li \textit{et. al} \cite{li2016deep} proposed a deep joint filtering (DJF) network, which used two independent convolutional neural networks (CNN) to extract the structural information from the guidance and input images, respectively, and then used a third CNN to predict the target image. Later, this work is further improved by adding a skip connection between the input and target images, which leads to DJFR \cite{li2019joint}. Kim \textit{et. al} \cite{kim2018deformable} proposed a deformable kernel network  (DKN) for joint image filtering, which employs a similar network architecture as DJF \cite{li2019joint}, but with a new added weight and offset learning module to learn the neighbourhood system for each pixel. Wu \textit{et. al} \cite{wu2018fast} turned the traditional guided filtering (GF) algorithm into a differentiable block, namely guided filtering layer, and then plug it into a convolutional neural network, so that it can be trained end-to-end. Recently, Pan \textit{et. al} \cite{pan2019spatially} proposed a spatially variant linear representation model, in which the target image is linearly represented by the guidance and input images. They then use a deep CNN to estimate the representation coefficients to restore the target image.  For the specific MIR task, methods \cite{song2016deep, guo2018hierarchical, song2018deeply, wen2019deep}  aim to upscale the depth image with guidance from the RGB image using deep neural networks. Methods \cite{lahoud2018multi, shi2018deep, lohit2019unrolled} aim to improve the resolution of the multi-spectral image with the assistance of either RGB or panchromatic images. Note that the methods aiming for a specific MIR task may not perform well on  other tasks, because either the filters or the network architectures have been  tailored to the specific characteristics of the image modalities under consideration. 
\vspace{-.5em}
\subsection{Multi-modal image fusion}
Multi-modal image fusion aims to  integrate the complementary information contained in   different source images to generate a fused image. The fused image should provide more comprehensive information about the scene, which is more helpful for human or machine perception. The source images are usually captured by different sensors, e.g., the MRI and CT medical images, but sometimes they are obtained using the same sensor but with different imaging parameters, e.g., multi-focus  and multi-exposure images.

Most traditional image fusion methods follow a three-step fusion procedure \cite{li2017pixel}. Firstly,   the source images are mapped into a specific transform domain, e.g., wavelet transform. Then, the transform coefficients are fused based on a fusion rule, and finally the fused coefficients are transformed back into the image domain to obtain the final fused image. There are two elements here which play a critical role in the fusion performance: the selection of the transform domain and the fusion rule.  Many works have studied  the fusion performance in different transform domains, including for example discrete wavelet transform (DWT) \cite{pajares2004wavelet}, discrete cosine transform (DCT) \cite{cao2014multi}, non-subsampled contourlet transform \cite{zhang2009multifocus}. There are also some papers based on sparse coding \cite{yang2012pixel,wei2015hyperspectral,liu2016image}, which fuse the sparse representations in the sparse domain. The most widely used fusion rules are choose-max \cite{hu2012multiscale} and weighted average \cite{jang2012contrast}. Usually, in order to avoid coefficient inconsistency, a neighbourhood morphological processing step is used after the choose-max strategy\cite{li2017pixel}.  However, the transform domain is manually selected and the fusion rule is hand-crafted, which may influence the fusion results.  

Recently, some authors have proposed to use deep learning to solve the MIF problems. Specifically, Liu \textit{et al.} \cite{liu2017multi}  proposed a simple CNN to predict the decision map for multi-focus image fusion. Prabhakar \textit{et al.} \cite{prabhakar2017deepfuse} proposed a CNN based unsupervised image fusion method to fuse one under-exposed image with an  over-exposed one. Li \textit{et al.}  \cite{li2018densefuse} proposed a CNN network with the dense block structure to solve the infrared and visible image fusion problem. To improve the perceptual quality of the fused image, Ma \cite{ma2019fusiongan} proposed a generative adversarial network (GAN), called FusionGAN, for infrared and visual image fusion.

The MIR and MIF tasks are usually treated as two independent research problems, as shown in the literature review above. One main contribution of this paper is that we propose  a novel multi-modal convolutional sparse coding (MCSC) model with two variations for these two tasks. Based on this model, we further propose a universal deep learning framework to achieve both the MIR and MIF tasks, as shown in Fig. \ref{frame}. To the best of our knowledge, this is the first time a general deep learning framework is proposed that can address either the MIR or the MIF problem.

\section{Multi-modal Convolutional Sparse Coding (MCSC)} \label{sparse}
In multi-modal image processing, one important issue is to model the dependencies among different modalities. In this section, we first introduce the multi-modal convolutional sparse coding (MCSC) model for the MIR and MIF tasks, and then introduce the optimization formulations, together with the solutions  for these two tasks. To make the notations clear, we list the main symbols used in this section  in Table \ref{nota}.

\begin{table} 	
	\caption{Summary of notations }\label{nota}	
	\centering
	\begin{tabular}{c|cc}
		\hline
		\hline			
		Notation&Dimension&Description\\ \hline
		$\bm{x}$ & $\mathds{R}^{n\times n}$&  the input image \\
		$\bm{y}$ & $\mathds{R}^{n\times n}$&  the guidance image \\
		$\bm{z}$ & $\mathds{R}^{n\times n}$&  the target image \\
		$\bm{d}_k^c$  & $\mathds{R}^{s\times s}$& the $k$-th common filter of  $\bm{x}$ \\
		$\bm{d}_k^u$  & $\mathds{R}^{s\times s}$& the $k$-th unique filter of  $\bm{x}$ \\
		$\bm{h}_k^c$  & $\mathds{R}^{s\times s}$& the $k$-th common filter of  $\bm{y}$ \\
		$\bm{h}_k^u$  & $\mathds{R}^{s\times s}$& the $k$-th unique filter of  $\bm{y}$ \\
		$\bm{g}_k^c$  & $\mathds{R}^{s\times s}$& the $k$-th common filter of  $\bm{z}$ \\
		$\bm{g}_k^u$  & $\mathds{R}^{s\times s}$& the $k$-th unique filter of  $\bm{z}$ related with $\bm{x}$ \\
		$\bm{g}_k^v$  & $\mathds{R}^{s\times s}$& the $k$-th unique filter of  $\bm{z}$ related with $\bm{y}$\\
		$\bm{c}_k$  & $\mathds{R}^{n\times n}$& the $k$-th common feature response  \\
		$\bm{u}_k$  & $\mathds{R}^{n\times n}$& the $k$-th unique feature response  of  $\bm{x}$\\
		$\bm{v}_k$  & $\mathds{R}^{n\times n}$& the $k$-th unique feature response  of  $\bm{y}$\\
		\hline				
	\end{tabular}	
\end{table}

\subsection{The MCSC model}
\textbf {For the MIR task,}
the problem can be formulated as follows: given a distorted image $\bm{x} \in \mathds{R}^{n\times n}$, and a guidance image $\bm{y}\in \mathds{R}^{n\times n}$,  we aim to reconstruct an image $\bm{z} \in\mathds{R}^{n\times n}$ which is a high-quality version of $\bm{x}$. Here, we assume that the image is square. Since  $\bm{x}$ and  $\bm{y}$ capture the same scene, they should share some common features, but they also have unique features. Take the depth and RGB images for example,   discontinuities in the depth image are clearly related to edges in RGB image. However,  RGB image  contains texture information, which has no relations with depth image. Based on this observation, 
we model the relationships among different modalities as follows:
\begin{equation} \label{ee1}
\bm{x}=\sum_k \bm{d}_k^c*\bm{c}_k+\sum_k \bm{d}_k^u*\bm{u}_k,
\end{equation}
\begin{equation} \label{ee1-1}
\bm{y}=\sum_k \bm{h}_k^c*\bm{c}_k+\sum_k \bm{h}_k^v*\bm{v}_k,
\end{equation}
\begin{equation} \label{ee1-2}
\bm{z}=\underbrace{\sum_k \bm{g}_k^c*\bm{c}_k}_{\text{\begin{tabular}{c}common features \\ between $\bm{x}$ and $\bm{y}$\end{tabular}}}+\underbrace{\sum_k \bm{g}_k^u*\bm{u}_k}_{\text{\begin{tabular}{c}unique features\\of $\bm{x}$\end{tabular}}}
\end{equation}
Here, the symbol * denotes the convolutional operation, $\{\bm{d}_k^c\}_{k=1}^K $, $\{\bm{h}_k^c\}_{k=1}^K $ and  $\{\bm{g}_k^c\}_{k=1}^K \in \mathds{R}^{s\times s \times K}$ are the common filters of $\bm{x}$, $\bm{y}$ and $\bm{z}$, respectively; $\{\bm{d}_k^u\}_{k=1}^K$ and $\{\bm{h}_k^v\}_{k=1}^K$  are the unique filters of $\bm{x}$ and $\bm{y}$, respectively; $\{\bm{g}_k^u\}_{k=1}^K \in \mathds{R}^{s\times s \times K}$ is the unique filter of $\bm{z}$ which is related with $\bm{x}$. The common filters share the same feature responses $\{\bm{c}_k\}_{k=1}^K\in \mathds{R}^{n\times n \times K}$, while the unique filters have their own feature responses $\{\bm{u}_k\}_{k=1}^K$ and  $\{\bm{v}_k\}_{k=1}^K\in \mathds{R}^{n\times n \times K}$. In summary, $\bm{x}$ and $\bm{y}$ share the same common feature responses $\{\bm{c}_k\}_{k=1}^K$, but also have different unique feature responses $\{\bm{u}_k\}_{k=1}^K$ and $\{\bm{v}_k\}_{k=1}^K$ as indicated in Eqs. \eqref{ee1} and \eqref{ee1-1}. Since  the unique information in the guidance image $\bm{y}$ is not useful for the reconstruction of $\bm{z}$, the model of $\bm{z}$ is composed of two parts: the features which are in common between  $\bm{x}$ and $\bm{y}$, and the unique features of $\bm{x}$, as in Eq. \eqref{ee1-2}. The advantage of this model is that only the useful information is preserved while the  information that may interfere with the estimation of $\bm{z}$ is discarded.

\textbf {For the MIF task,}
we aim to fuse the image $\bm{x}$ with another image $\bm{y}$ to obtain a new image $\bm{z}$ which has both the advantages of $\bm{x}$ and $\bm{y}$. Different from the MIR task in which only the common information in $\bm{y}$ is helpful,  here in the MIF task, all the information contained in $\bm{y}$ might be useful for the reconstruction of $\bm{z}$. 	
In other words, the fused image $\bm{z}$ should be composed of three parts: common features shared by  $\bm{x}$ and  $\bm{y}$, unique features of  $\bm{x}$ and unique features of  $\bm{y}$.  Thus, in the MIF task, the models for $\bm{x}$ and $\bm{y}$ are the same as for the MIR task, but the model for  $\bm{z}$ should be re-formulated as follows:

\begin{equation} \label{ee2}
\bm{z}=\underbrace{\sum_k \bm{g}_k^c*\bm{c}_k}_{\text{\begin{tabular}{c}common features \\ between $\bm{x}$ and $\bm{y}$\end{tabular}}}+\underbrace{\sum_k \bm{g}_k^u*\bm{u}_k}_{\text{\begin{tabular}{c}unique features\\of $\bm{x}$\end{tabular}}}+ \underbrace{\sum_k \bm{g}_k^v*\bm{v}_k}_{\text{\begin{tabular}{c}unique features\\of $\bm{y}$\end{tabular}}}, 
\end{equation}
where $\{\bm{g}_k^u\}_{k=1}^K$ and $\{\bm{g}_k^v\}_{k=1}^K\in \mathds{R}^{s\times s\times K}$  are the unique filters of $\bm{z}$ related with $\bm{x}$ and $\bm{y}$, respectively.

\subsection{Convolutional Sparse Coding } \label{rerew}
In the synthesis phase, given the input images $\bm{x}$ and $\bm{y}$, we need to infer the image $\bm{z}$. To this end, assuming all dictionaries are known, we  need to calculate the common and unique feature responses $\{\bm{c}_k, \bm{u}_k, \bm{v}_k\}_{k=1}^K$, through solving the following optimization problem:

\begin{equation} \label{ee3}
\begin{split}
\underset{\begin{tiny}\{\begin{matrix} \bm{c}_k,\bm{u}_k,\bm{v}_k \end{matrix}\}\end{tiny}}{\mathrm{Argmin}} &\frac{1}{2}\left\|\bm{x}-\sum_k(\bm{d}_k^c*\bm{c}_k+\bm{d}_k^u*\bm{u}_k) \right\|_2^2\\+&\frac{1}{2}\left\|\bm{y}-\sum_k(\bm{h}_k^c*\bm{c}_k+\bm{h}_k^v*\bm{v}_k) \right\|_2^2\\
+&\lambda \sum_k(\left\|\bm{c}_k\right\|_1+\left\|\bm{u}_k\right\|_1+\left\|\bm{v}_k\right\|_1).
\end{split}
\end{equation}
 To solve this problem, our strategy is to alternately update each variable with other variables fixed.  This leads to the following three steps:

\begin{itemize}
	\item  Step 1, we  fix the common feature response $\bm{c}_k$ and the  unique response $\bm{v}_k$ of $\bm{y}$, to update the  unique response $\bm{u}_k$ of $\bm{x}$. 
	\item Step 2, we fix the common feature response $\bm{c}_k$ and the  unique response $\bm{u}_k$ of $\bm{x}$, to update the  unique response $\bm{v}_k$ of $\bm{y}$.  
	\item Step 3, we fix all the unique responses $\bm{u}_k$ and $\bm{v}_k$ to update the common response $\bm{c}_k$. 
\end{itemize}

These three steps should be repeated until convergence. After solving \eqref{ee3}, the target image $\bm{z}$ can be reconstructed by either Eq. \eqref{ee1-2} or Eq. \eqref{ee2}, depending on the task. However, solving \eqref{ee3} is very time-consuming, since the solution requires several iterations. In this paper, we aim to solve \eqref{ee3} using a deep learning approach, by turning each step above into a deep network module with tunable parameters and adjustable number of layers. The next section introduces in detail how we turn these three steps into a deep network.

\section{ Common and Unique Information Splitting Network (CU-Net) } \label{dcus}
The architecture of the proposed CU-Net is shown in Fig. \ref{frame}. We can see that the network is composed of three parts: the unique feature extraction module (UFEM), common feature preservation module (CFPM), and the image reconstruction module (IRM). Next, we will introduce these three modules in detail and explain how we design the architecture of each of them from the MCSC model.
\subsection{Unique feature extraction module (UFEM)}
The UFEM aims to extract the unique features from each source image. Since we have two source images $\bm{x}$ and $\bm{y}$,  we have two UFEMs, i.e., the $\{\bm{u}_k\}$ prediction network and the $\{\bm{v}_k\}$ prediction network, as shown in Fig. \ref{frame}. As discussed in Section \ref{rerew}, the first step is to fix $\bm{c}_k$ and $\bm{v}_k$ to update $\bm{u}_k$, which turns \eqref{ee3} into the following:
\begin{equation} \label{ee4}
\underset{\begin{tiny}\{ \bm{u}_k \}\end{tiny}}{\mathrm{Argmin}} \frac{1}{2}\left\|\bm{x}-\sum_k(\bm{d}_k^c*\bm{c}_k+\bm{d}_k^u*\bm{u}_k) \right\|_2^2
+\lambda \sum_k\left\|\bm{u}_k\right\|_1.
\end{equation}
Since $\bm{c}_k$ is fixed, we can further simplify \eqref{ee4} by denoting $\bm{\hat{x}}=\bm{x}-\sum_k \bm{d}_k^c*\bm{c}_k$, which leads to the following expression:
\begin{equation} \label{ee5}
\underset{\begin{tiny}\{ \bm{u}_k \}\end{tiny}}{\mathrm{Argmin}} \frac{1}{2}\left\|\bm{\hat{x}}-\sum_k \bm{d}_k^u*\bm{u}_k \right\|_2^2
+\lambda \sum_k\left\|\bm{u}_k\right\|_1.
\end{equation}
Eq. \eqref{ee5} is a standard convolutional sparse coding problem, which can be solved using the traditional method in \cite{wohlberg2015efficient}. We instead solve this problem using the learned convolutional sparse coding (LCSC) algorithm proposed in  \cite{sreter2018learned}, 
which gives us the solution of $\bm{u}_k$ as follows:
\begin{equation} \label{ee6}
\bm{U}_{j+1}=S_{\lambda}(\bm{U}_j-\bm{E}_{u}*\bm{D}_{u}*\bm{U}_j+\bm{E}_{u}*\bm{\hat{x}}),
\end{equation}
where $\bm{U}\in \mathds{R}^{n\times n\times K }$ is the stack of feature responses $\{\bm{u}^k\}_{k=1}^K$, and  $\bm{U}_j$ is the update of $\bm{U}$ at the $j$-th iteration, $\bm{D}_{u} \in \mathds{R}^{s\times s\times K}$ and  $\bm{E}_{u} \in \mathds{R}^{s\times s\times K}$ are the learnable convolutional layers  related to the filters $\{\bm{d}_k^u\}_{k=1}^K$.  For details  about how to derive \eqref{ee6} by solving \eqref{ee5}, please refer to the Appendix A.  The  $S_{\lambda} (\cdot)$ indicates the soft-thresholding operation with $\lambda$ as the threshold. Following \cite{sreter2018learned}, we can unfold the iterations in \eqref{ee6} into a neural network, as shown in the UFEM module (the upper figure) in Fig. \ref{module}. Theoretically, since each LCSC block in the UFEM module corresponds to one iteration in \eqref{ee6}, the UFEM module can be extended to any number of LCSC blocks, which makes the network architecture very flexible. 

The architecture of  $\{\bm{v}_k\}$ prediction network is obtained in a way similar  to that of $\{\bm{u}_k\}$, and it is derived from the second step in solving  Eq. \eqref{ee3}. When updating $\bm{v}_k$, we need to fix $\bm{c}_k$ and $\bm{u}_k$, so that Eq. \eqref{ee3} becomes:
\begin{equation} \label{ee7}
\underset{\begin{tiny}\{ \bm{v}_k \}\end{tiny}}{\mathrm{Argmin}} \frac{1}{2}\left\|\bm{\hat{y}}-\sum_k \bm{h}_k^v*\bm{v}_k \right\|_2^2
+\lambda \sum_k\left\|\bm{v}_k\right\|_1,
\end{equation}
where $\bm{\hat{y}}=\bm{y}-\sum_k \bm{h}_k^c*\bm{c}_k$. Likewise, we can use the LCSC algorithm from \cite{sreter2018learned} to solve \eqref{ee7}, which gives us the following iterations:
\begin{equation} \label{ee8}
\bm{V}_{j+1}=S_{\lambda}(\bm{V}_j-\bm{E}_{v}*\bm{D}_{v}*\bm{V}_j+\bm{E}_{v}*\bm{\hat{y}}).
\end{equation}
Here, $\bm{V}\in \mathds{R}^{n\times n\times K}$ is the stack of feature responses $\{\bm{v}^k\}_{k=1}^K$, and  $\bm{V}_j$ is the update of $\bm{V}$ at the $j$-th iteration, $\bm{D}_{v} \in \mathds{R}^{s\times s\times K}$ and $\bm{E}_{v} \in \mathds{R}^{s\times s\times K}$ are the learnable convolutional layers related with $\{\bm{h}_k^v\}_{k=1}^K$. The iterations in Eq. \eqref{ee8} can be also unfolded into a neural network, as shown in the middle figure in Fig. \ref{module}.

\subsection{Common feature preservation module (CFPM)}
Since different modalities capture the same image scene, there exist some consistent  features among them.  The CFPM aims to  preserve these common features. The network architecture of CFPM is derived from the third step in solving Eq. \eqref{ee3} which aims to predict the common features $\{c_k\}$. Thus, the CFPM is also called the $\{c_k\}$ prediction network in this paper. When fixing the unique responses  $u_k$ and $v_k$, Eq. \eqref{ee3} can be re-written as follows:
\begin{equation} \label{ee9}
\begin{split}
\underset{\begin{tiny}\{ \bm{c}_k \}\end{tiny}}{\mathrm{Argmin}} &\frac{1}{2}\left\|\bm{\tilde{x}}-\sum_k \bm{d}_k^c*\bm{c}_k \right\|_2^2+\frac{1}{2}\left\|\bm{\tilde{y}}-\sum_k \bm{h}_k^c*\bm{c}_k \right\|_2^2
\\+&\lambda \sum_k\left\|\bm{c}_k\right\|_1,
\end{split}
\end{equation}
where $\bm{\tilde{x}}=\bm{x}-\sum_k \bm{d}_k^u*\bm{u}_k$ and $\bm{\tilde{y}}=\bm{y}-\sum_k \bm{h}_k^v*\bm{v}_k$. The first two terms in Eq. \eqref{ee9} can be further combined and this yields the following optimization problem:
\begin{equation} \label{ee10}
\underset{\begin{tiny}\{ \bm{c}_k \}\end{tiny}}{\mathrm{Argmin}} \frac{1}{2}\left\|\bm{p}-\sum_k \bm{l}_k^c*\bm{c}_k \right\|_2^2
+\lambda \sum_k\left\|\bm{c}_k\right\|_1,
\end{equation}
where $\bm{p}\in \mathds{R}^{n\times n\times 2}$ is the concatenation of $\bm{\tilde{x}}$ and $\bm{\tilde{y}}$, and $\bm{l}_k^c \in\mathds{R}^{s\times s\times 2}$ is the concatenation of $\bm{d}_k^c$ and $\bm{h}_k^c$.
 Eq. \eqref{ee10} can be then solved using again the LCSC algorithm from \cite{sreter2018learned}. This leads to the following solution:
\begin{equation} \label{ee11}
\bm{C}_{j+1}=S_{\lambda}(\bm{C}_j-\bm{E}_{c}*\bm{D}_{c}*\bm{C}_j+\bm{E}_{c}*\bm{p}),
\end{equation}
where $\bm{C}\in \mathds{R}^{n\times n\times K}$ is the stack of common feature responses $\{\bm{c}^k\}_{k=1}^K$ and  $\bm{C}_j$ is the update of $\bm{C}$ at the $j$-th iteration, $\bm{D}_{c} \in \mathds{R}^{s\times s\times K\times2}$ and $\bm{E}_{c} \in \mathds{R}^{s\times s\times 2\times K}$ are the learnable convolutional layers related with  $\{\bm{l}_k^c\}_{k=1}^K$.
The iterations in Eq. \eqref{ee11} can be unfolded and this leads to the CFPM architecture shown in the bottom figure in Fig. \ref{module}.
\begin{table*} \addtolength{\tabcolsep}{-5pt}
	\small
	\caption{Results on the RGB/depth  dataset for 4$\times$ upscaling, with the best results in bold and the second best results underlined. }\label{NT}
	\vspace{-1em}
	\begin{center}
		\begin{tabular}{c|cccccccccccccccc}
			\hline
			\hline
			RGB/depth& \multicolumn {2}{c}{Ambush}&\multicolumn{2}{c}{Cave}&\multicolumn{2}{c}{Market}&\multicolumn{2}{c}{Art}&\multicolumn{2}{c}{Books}&\multicolumn{2}{c}{Moebius} &\multicolumn{2}{c}{Reindeer}&\multicolumn {2}{c}{Average}\\ \hline
			Methods&RMSE&SSIM&RMSE&SSIM&RMSE&SSIM&RMSE&SSIM&RMSE&SSIM&RMSE&SSIM&RMSE&SSIM&RMSE&SSIM\\
			Bicubic                                    &6.39 &0.9685  &6.61 &0.9503 &8.83 &0.9295 & 3.87&0.9687 &1.60 &0.9911 &1.32 &0.9908&2.81&0.9852&4.49&0.9692\\
			Xie \textit{et al.} \cite{xie2016edge}       &8.79 &0.9438 &9.14 &0.9221 &12.21& 0.8869 &3.79 &0.9758 &1.63 &0.9917 &1.33 &0.9910&2.77&0.9887&5.67&0.9571\\								
			Park \textit{et al.} \cite{park2011high}     &6.03 &0.9678 &7.13 &0.9379 &9.45 &0.9067 &3.76 &0.9752 &1.66 &0.9912 &1.42 &0.9911&2.79&0.9864&4.61&0.9638\\
			Ferstl \textit{et al.} \cite{ferstl2015}&5.99 &0.9701  &6.40 &0.9563 &8.01 & 0.9298 &3.73 &0.9771 &1.65 &0.9915 &1.43 &0.9909&2.90&0.9859&4.30&0.9717\\
			Lu \textit{et al.} \cite{lu2015sparse}       &5.53 &0.9712  &6.10 &0.9610 &8.31& 0.9266 &4.10 &0.9747 &2.18 &0.9896 &1.56 &0.9896&3.24&0.9867&4.43&0.9713\\
			Gu \textit{et al.}\cite{gu2017learning}    &6.04&0.9766&6.15&0.9613&8.10&0.9470 &3.52 &0.9779 &1.57 &0.9923 &1.23 &0.9930&2.66&0.9883&4.18&0.9766\\		   
			SCN\cite{liu2016robust} &4.29&0.9850&4.37&0.9769&5.94&0.9664&2.59 &0.9858 &1.08 &0.9951 &0.93 &0.9949&1.94&0.9921&3.02&0.9852\\			
			EDSR \cite{lim2017enhanced}  &2.90&0.9932&3.10&0.9890&4.64&0.9823&1.82&0.9937&0.64&0.9977&0.68&0.9970&1.27&0.9962&2.13&0.9929\\	
			SRFBN \cite{li2019feedback}           &2.96&0.9931&3.13&0.9887&4.55&0.9823&1.83&0.9936&0.67&0.9976&0.70&0.9969&1.30&0.9959&2.42&0.9929\\	
			RCGD\cite{liu2017robust}&5.62 & 0.9818 &5.92 &0.9647 &7.62 &0.9574 &3.06 &0.9834 &1.41 &0.9930 &1.19 &0.9927 &2.52&0.9894&3.90&0.9803\\
			RADAR \cite{deng2019radar}&3.43&0.9919 &3.57 &0.9857 &5.13 &0.9782 &2.00 &0.9922 &0.68 &0.9978 &0.73 &0.9969 &1.43&0.9956&2.42&0.9912\\ 
			
			DGF\cite{wu2018fast}&16.57&0.8914&18.09&0.7623&18.66&0.7877&10.73&0.9113&5.25&0.9712&4.70&0.9630&6.65&0.9479&11.52&0.8907\\	DJFR\cite{li2019joint} &2.95&0.9920&3.10&0.9889&4.56&0.9814&1.96&0.9928&0.68&0.9977&0.66&0.9970&1.15&0.9968&2.15&0.9924\\
			CoISTA\cite{deng2019deep} &\underline{2.71}&\underline{0.9939}&\underline{2.94}&\underline{0.9885}&\underline{4.42}&\underline{0.9806}&\underline{1.61}&\underline{0.9948}&\underline{0.59}&\underline{0.9977}&\underline{0.65}&\underline{0.9970}&\underline{1.14}&\underline{0.9969}&\underline{2.01}&\underline{0.9928}\\										
			CU-Net & \textbf{2.52}&\textbf{0.9947} &\textbf{2.75} &\textbf{0.9895} &\textbf{4.09} &\textbf{0.9832}  &\textbf{1.54} &\textbf{0.9954} &\textbf{0.55} &\textbf{0.9980} &\textbf{0.63} &\textbf{0.9973}&\textbf{1.10}&\textbf{0.9972}&\textbf{1.88}&\textbf{0.9936}\\		
			\hline								
		\end{tabular}
	\end{center}
\end{table*}
\subsection{Image reconstruction module (IRM)}
After  we  obtain the feature responses $\{c_k, u_k, v_k\}_{k=1}^K$,  the next step is to reconstruct the target image $\bm{z}$ using either Eq. \eqref{ee1-2} for MIR related tasks or Eq. \eqref{ee2} for MIF related tasks. Specifically, for MIR related tasks, the image reconstruction module (IRM) only contains two set of filters $\{\bm{g}_k^u\}_{k=1}^K$ and $\{\bm{g}_k^c\}_{k=1}^K$. As shown in Fig. \ref{frame}, the filter $\{\bm{g}_k^u\}_{k=1}^K$ is directly connected to the output of $\{u_k\}$ prediction network, which  corresponds to the term $\sum_k \bm{g}_k^u*\bm{u}_k$ in Eq. \eqref{ee1-2}.  The filter $\{\bm{g}_k^c\}_{k=1}^K$ is  connected to the output of $\{c_k\}$ prediction network, which  corresponds to the term $\sum_k \bm{g}_k^c*\bm{c}_k$ in Eq. \eqref{ee1-2}. Then, these two terms are added to get $\bm{z}$. For the MIF related tasks, we use Eq. \eqref{ee2} to create the IRM, and it contains three set of filters $\{\bm{g}_k^u\}_{k=1}^K$, $\{\bm{g}_k^v\}_{k=1}^K$, and  $\{\bm{g}_k^c\}_{k=1}^K$.  The only difference to the MIR related task is the filter $\{\bm{g}_k^v\}_{k=1}^K$, which is connected to the output of $\{v_k\}$ prediction network, as shown in Fig. \ref{frame}.
\subsection{Discussion about the CU-Net Architecture}
It is of interest to note that, although the CU-Net is derived from a theoretical MCSC model, it contains both skip connections and residual  blocks. These two elements have both been demonstrated to improve the reconstruction performance of CNN architectures for different tasks.  The concept of skip connection was first proposed by He \textit{et al.} \cite{he2016deep} for image recognition, and later successfully used in the field of image super-resolution \cite{kim2016accurate,lai2017deep} and image denoising \cite{zhang2017beyond}. In our network, as shown in Fig. \ref{frame}, the target $\bm{z}$ is obtained by combining three parts, in which the two parts convolved with $\bm{g}_k^u$ and $\bm{g}_k^v$ both use the skip connections.
In addition, as we can see in Fig. \ref{module}, there are two residual lines in our UFEM and CFPM,  which are  similar to the residual  block proposed in \cite{lim2017enhanced}. As verified in \cite{lim2017enhanced}, the residual block can make full use of the hierarchical features and thus significantly improve the image super-resolution performance. In this paper, we also demonstrate the effectiveness of the residual UFEM and CFPM architecture in Section \ref{abh}. 

\section{Experiments} \label{expe}
In this section, we verify the effectiveness of our method on various applications, which can be classified into two categories:  MIR  and MIF related tasks.  The MIR related tasks include  RGB guided depth image super-resolution, RGB guided multispectral image super-resolution, and flash guided non-flash image denoising. The MIF related tasks include  multi-exposure image fusion, multi-focus image fusion and medical image fusion.  

\textbf{Training details.} For each task, we train the network using around 150,000 image patches with size $64\times64$. The Adam optimizer is used to train the network, with a basic learning rate of $1e^{-4} $ and it is decayed by 0.9 every 50 epochs. The number of iterated LCSC blocks in the UFEM and CFPM modules is set to 4. For each convolutional layer, the filter size is $8\times8$, and the number of filters is 64. We choose the size of mini-batch as 64 and the total number of epochs as 200. Note that for applications which aim to restore an image with multiple channels, e.g., the multi-exposure image fusion, instead of processing each channel independently, we adjust the dimensions of corresponding convolutional layers to accept multi-channel inputs. Specifically, we have $\bm{D}_u$ and $\bm{D}_v \in \mathds{R}^{s\times s\times K \times m}$,   $\bm{E}_u$ and $\bm{E}_v \in \mathds{R}^{s\times s\times m \times K} $, $\bm{D}_c \in \mathds{R}^{s\times s\times K \times 2m} $ and $\bm{E}_c \in \mathds{R}^{s\times s\times 2m \times K} $, where $m$ is the number of channels.

Next, in Section \ref{mrt}, we show the simulation results of our method on the MIR related tasks, together with the comparisons with other state-of-the-art approaches. In Section 5.2, we show the simulation results on the MIF related tasks. In order to further verify the effectiveness of our deep network, we visualize in Section 5.3 the features generated by different parts of the network, and in Section 5.4 we present  comprehensive ablation study results. In Section 5.5, the computational cost is discussed.

%
\begin{table*} \addtolength{\tabcolsep}{-4pt}
	\small
	\caption{Results on the RGB/MS  dataset for 4$\times$ upscaling, with the best results in bold and the second best results underlined. }\label{MS}
	\vspace{-1em}
	\begin{center}
		\begin{tabular}{c|cccccccccccccccc}
			\hline
			\hline
			RGB/MS& \multicolumn {2}{c}{Chart toy}&\multicolumn{2}{c}{Egyptian}&\multicolumn{2}{c}{Feathers}&\multicolumn{2}{c}{Glass tiles}&\multicolumn{2}{c}{Jelly beans}&\multicolumn{2}{c}{Oil painting
			}&\multicolumn {2}{c}{Paints}& \multicolumn {2}{c}{Average}\\ \hline
			Methods&PSNR&SSIM&PSNR&SSIM&PSNR&SSIM&PSNR&SSIM&PSNR&SSIM&PSNR&SSIM&PSNR&SSIM&PSNR&SSIM\\
			Bicubic                                    &28.94 &0.9424 &36.57 &0.9786 &30.80 &0.9562 &26.65 &0.9242 &27.81&0.9302 &31.67&0.8943 &29.29 &0.9493&30.25&0.9393\\
			
			JBF\cite{kopf2007joint}                    &32.56 &0.9653 &38.73 &0.9735 &33.60 &0.9637 &27.52 &0.9341 &30.29&0.9498 &32.77&0.8962 &31.94 &0.9699&32.49&0.9504\\								
			GF\cite{he2010guided}                      &34.09 &0.9788 &40.24 &0.9796 &33.60 &0.9748 &29.46 &0.9593 &30.90&0.9658 &35.03&0.9441 &31.73 &0.9702&33.58&0.9675\\
			SDF\cite{ham2018robust}                    &31.87 &0.9694 &39.43 &0.9795 &33.45 &0.9650 &28.22 & 0.9374&30.32&0.9433 &32.86&0.9126 &31.96 &0.9655&32.59&0.9532\\			
			JFSM\cite{shen2015multispectral}           &32.98 &0.9295 &40.39 &0.9705 &33.89 &0.9425 &28.98 & 0.9397&31.18&0.9451 &35.91&0.9560 &32.76 &0.9430&33.73&0.9466\\			
			
			EDSR \cite{lim2017enhanced}           &33.45&0.9836&40.03&0.9829&35.55&0.9875&29.75&0.9706&32.81&0.9838&32.69&0.9178&37.28&0.9914&34.51&0.9739\\
			SRFBN \cite{li2019feedback} &33.43&0.9838&40.04&0.9822&35.53&0.9873&29.53&0.9676&32.97&0.9845&32.68&0.9182&36.06&0.9907&34.32&0.9735\\
			MMSR \cite{lahoud2018multi}	&37.55&0.9934&45.21&0.9919&39.68&0.9933&34.52&0.9912&38.87&0.9883&37.50&0.9797&39.31&0.9943&38.95 & 0.9903  \\ 	
			DGF\cite{wu2018fast}&34.19&0.9559&37.81&0.9620&31.22&0.9336&29.93&0.9339&28.94&0.9459&36.10&0.9649&31.72&0.9680&32.85&0.9520\\
			DJFR\cite{li2019joint}                       &\underline{37.86} &\underline{0.9935}&45.69 &0.9922 &\underline{40.13} &\underline{0.9939} &\underline{34.97}& \underline{0.9915} &\underline{39.16} &\underline{0.9885} &\underline{37.76} &\underline{0.9805} &\underline{39.36} &\underline{0.9944}&\underline{39.28}&\underline{0.9906}\\
			CoISTA\cite{deng2019deep} &36.58&0.9914&\underline{45.91}&\textbf{0.9961}&39.62&0.9937&33.99&0.9907&38.92&0.9956&37.26&0.9690&38.40&0.9949&38.67&0.9902\\				
			CU-Net  &\textbf{39.65}&\textbf{0.9960} &\textbf{47.35}&\underline{0.9930}&\textbf{42.47}&\textbf{0.9965}&\textbf{35.93}&\textbf{0.9944}&\textbf{40.69}&\textbf{0.9972}&\textbf{38.84}&\textbf{0.9811}&\textbf{40.86}&\textbf{0.9971}&\textbf{40.83}&\textbf{0.9936}\\			
			\hline								
		\end{tabular}
	\end{center}
\vspace{-.5em}
\end{table*}
\subsection{MIR related tasks} \label{mrt}
\subsubsection{RGB guided depth image SR}
We use the  dataset  from \cite{riegler2016deep} to train the network, which is composed of 1000 synthetic RGB/Depth image pairs. The testing images are from the Middlebury dataset \cite{hirs2007} and the Sintel dataset \cite{butler2012naturalistic}. The upscaling factor is 4, and we generate the low-resolution depth image by downsampling the high-resolution depth image by the upscaling factor, and then upsample it  using the bicubic interpolation. For the high-resolution RGB  image, we convert it into the YCbCr format and only the Y channel is used as guidance.  

We compare our method with several state-of-the-art methods, which include methods developed for single image super-resolution, e.g.,  SCN\cite{liu2016robust}, EDSR \cite{lim2017enhanced}, SRFBN \cite{li2019feedback},   methods  for single depth image super-resolution and RGB guided depth image super-resolution, e.g., Xie \textit{et al.} \cite{xie2016edge}, Park \textit{et al.} \cite{park2011high}, Ferstl \textit{et al.} \cite{ferstl2015}, Gu \textit{et al.}\cite{gu2017learning}, RCGD\cite{liu2017robust}, RADAR \cite{deng2019radar}, and  methods for solving the general MIR problem, e.g., DGF\cite{wu2018fast}, DJFR\cite{li2019joint} and CoISTA\cite{deng2019deep}. The results of these approaches are all obtained by running the software codes provided by the authors.

\begin{figure*}
	\centering
	\centerline{\epsfig{figure=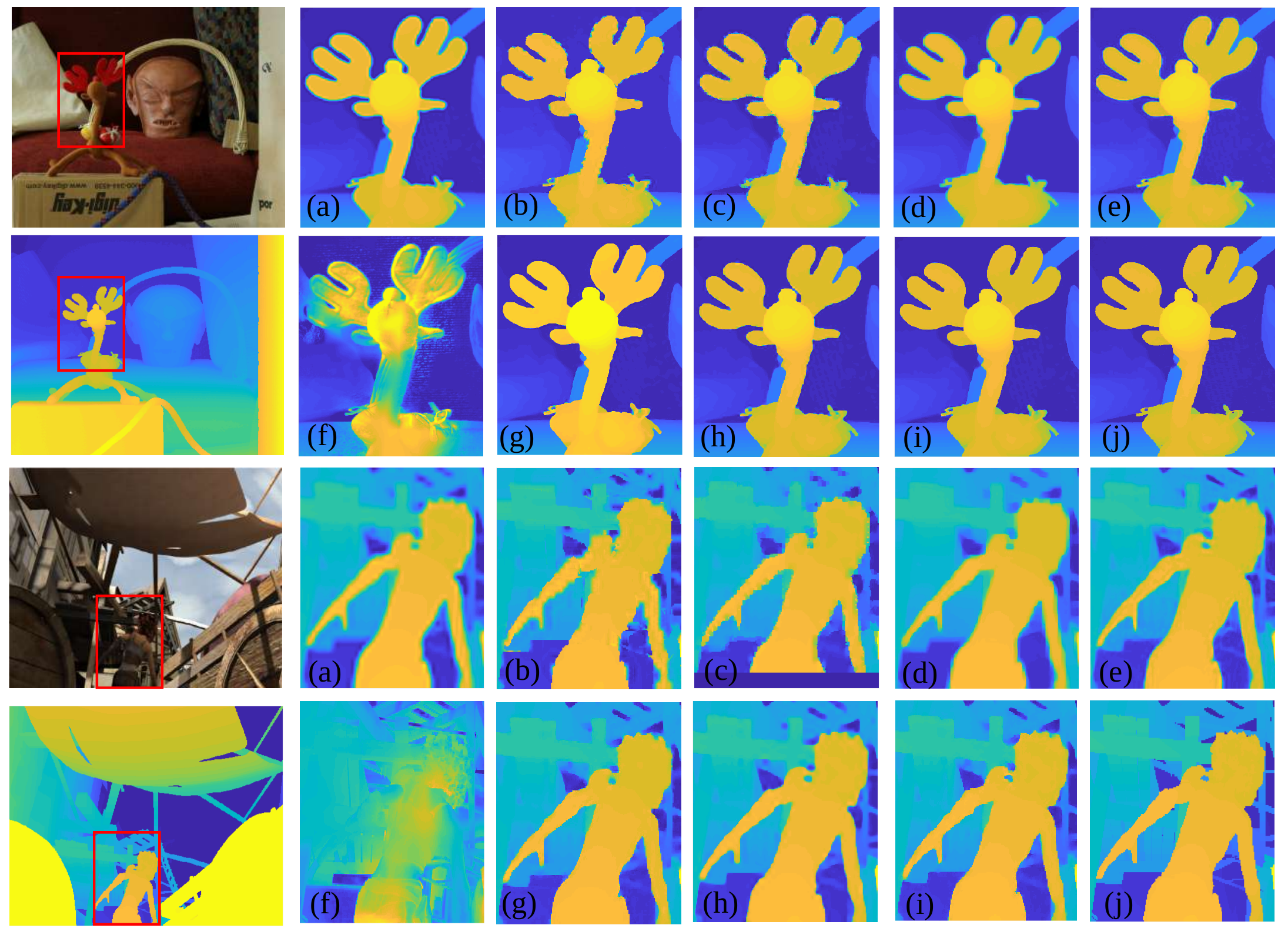, width=16cm}}
	\caption{Visual comparisons of RGB guided depth SR results for 4$\times$ upscaling: (a)  Bicubic, (b) Lu \textit{et al.} \cite{lu2015sparse}, (c) Xie \textit{et al.} \cite{xie2016edge}, (d) Gu \textit{et al.}\cite{gu2017learning}, (e) SCN\cite{liu2016robust}, (f) DGF\cite{wu2018fast}, (g)  DJFR\cite{li2019joint}, (h) RADAR \cite{deng2019radar}, (i) Our CU-Net, (j) Ground truth.  }\label{depth}
\end{figure*}

Table \ref{NT} presents the quantitative comparison results between ours and the other state-of-the-art methods, in terms of root mean square error (RMSE) and SSIM \cite{wang2004image}. As can be seen from this table,  our CU-Net achieves the best results among all the methods. Fig. \ref{depth} visualizes the reconstructed depth images using different methods for 4$\times$ upscaling. From this figure, we can see that  the depth images recovered by our method are quite close to the ground-truth, with clearer and sharper edges.
\subsubsection{RGB guided multi-spectral image SR}
\begin{figure*}
	\centering
	\centerline{\epsfig{figure=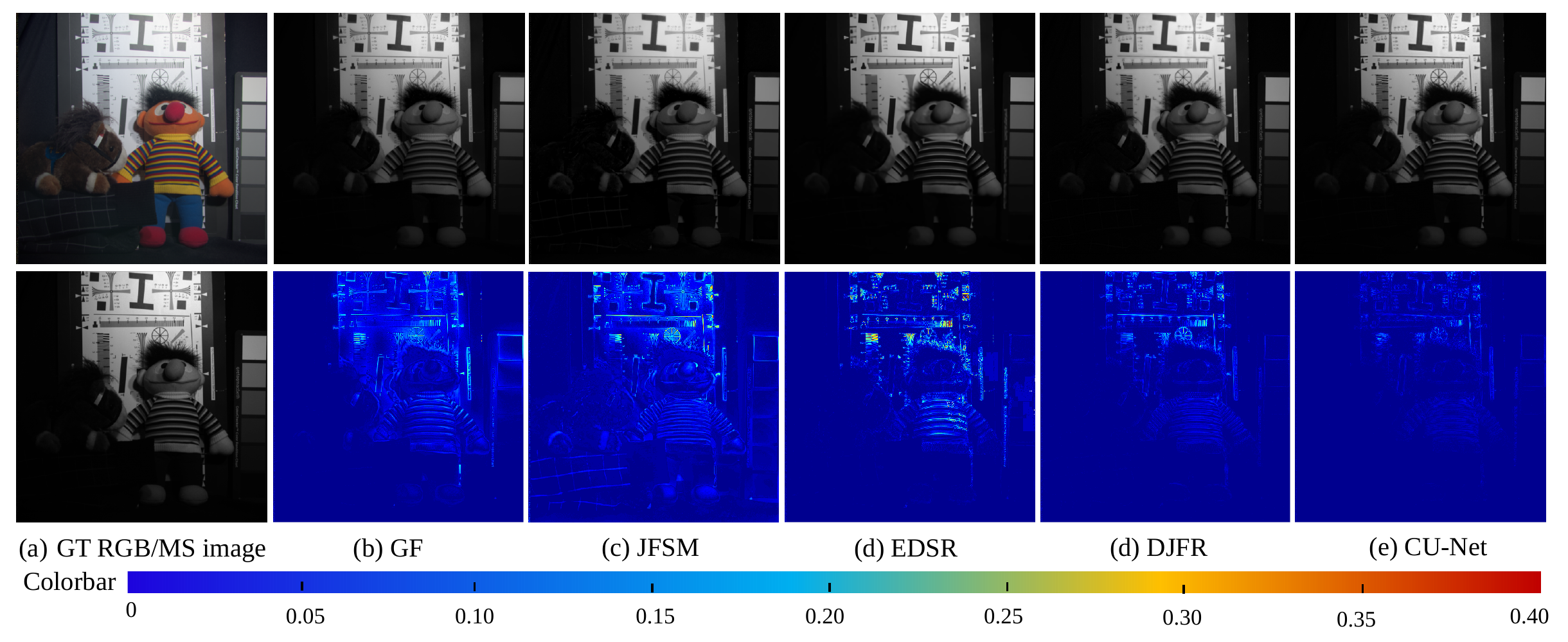, width=16.5cm}}
	\vspace{-1.5em}
	\caption{Visual comparisons of RGB guided multi-spectral image SR results for 4$\times$ upscaling. The first row shows the super-resolved images using different methods, and the second row shows the error maps between the images in the first row and the ground-truth.   }\label{mss}
\end{figure*}

For the task of RGB guided Multi-spectral image SR, both the training and testing images are from the the Columbia multi-spectral database \cite{yasuma2010}. We randomly select 7 images for testing and use the remaining  images for training.   To verify the effectiveness of our method, we compare it with the following approaches: JBF\cite{kopf2007joint}, GF\cite{he2010guided}, JFSM\cite{shen2015multispectral}, SDF\cite{ham2018robust}, EDSR \cite{lim2017enhanced}, SRFBN \cite{li2019feedback}, MMSR \cite{lahoud2018multi}, DGF\cite{wu2018fast},   DJFR\cite{li2019joint} and CoISTA\cite{deng2019deep}. Table \ref{MS} shows the comparison results in terms of PSNR and SSIM for 4$\times$ upscaling.  As we can see, our method can recover the HR multi-spectral images with high accuracy. Specifically,  the average PSNR value of our method is 1.55 dB higher than  the second best method DJFR\cite{li2019joint}. Fig. \ref{mss} shows examples of the upsampling results using different methods. 
\vspace{-0.3em}
\subsubsection{Flash guided non-flash image denoising}
We use the new flash/non-flash image dataset provided by \cite{aksoy2018dataset} for both training and testing.  We randomly select 400 flash/non-flash image pairs from the dataset \cite{aksoy2018dataset} for training.  For testing, to make the results more convincing, we select 12 images from three different categories in \cite{aksoy2018dataset}, i.e., the toy, plant and object. Note that the testing images are different from the training images.  The noisy non-flash images are obtained by adding white Gaussian noise to the clean images, with three different noise levels, i.e., $\sigma^2=25$, $50$, and $75$. 

\begin{figure*}
	\centering
	\centerline{\epsfig{figure=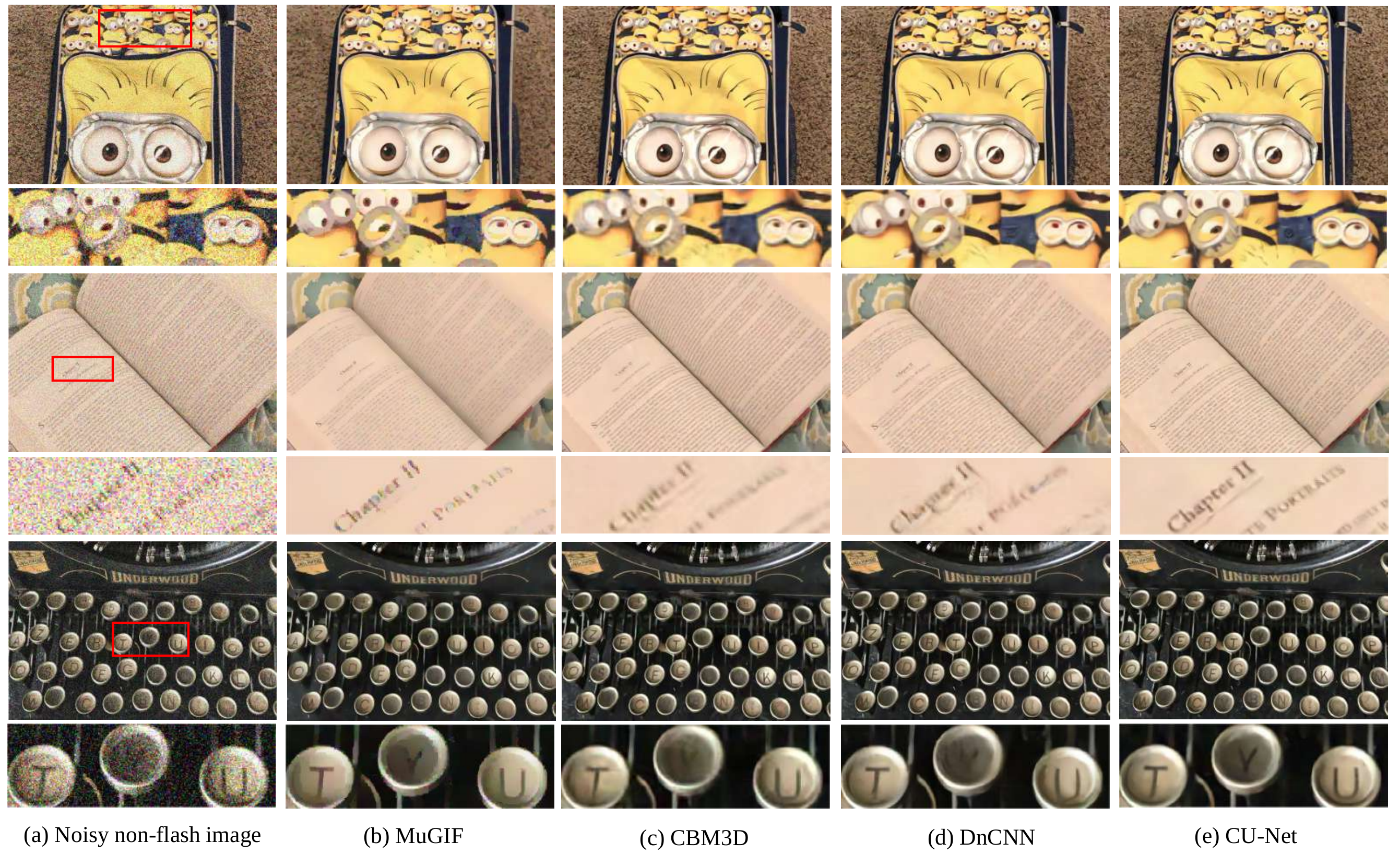, width=16cm}}
	\caption{Visual comparisons of flash guided non-flash image denoising results with $\sigma^2=75$: (a)  Noisy non-flash input image, (b) MuGIF \cite{guo2018mutually}, (c) CBM3D \cite{dabov2008image}, (d) DnCNN \cite{zhang2017beyond}, (e) Our CU-Net.  }\label{fla_noi}
\vspace{-1em}
\end{figure*}

To verify the effectiveness of our method, we compare it with the other four state-of-the-art methods, including CBM3D \cite{dabov2008image} and DnCNN  \cite{zhang2017beyond} which are specifically aimed at the image denoising task, DJFR\cite{li2019joint} and MuGIF \cite{guo2018mutually} which are general MIR methods.  Table \ref{flash} shows the comparison results in terms of PSNR, in which the first four images are from the toy category, the middle four images are from the object category and the last four images are from the plant category. As we can see from this table, our method outperforms other methods for different noise levels.  This  is further verified in Fig. \ref{fla_noi} in which we visualize the denoised non-flash images using different methods with $\sigma^2=75$.  In the case of $\sigma^2=75$, the non-flash image is severely affected by noise, with most of the details removed by the heavy noise, and thus the guidance of the flash image becomes very important. 
As we can see in Fig. \ref{fla_noi}, our method is able to remove the noise effectively, and at the same time recover the fine details and sharp edges, while others either have blurred edges \cite{dabov2008image,zhang2017beyond} or unclear details \cite{guo2018mutually}. Take the image book for example, we can clearly recover the texts which is hard to read in the noisy image, while the other methods struggle to do so. This confirms that our method is able to make full use of the information in the guidance image, even in the case that most  information in the noisy non-flash image has been lost.

\begin{table*} \addtolength{\tabcolsep}{-4pt}
	\small
	\caption{Results on the Flash/non-flash  dataset for image denoising, with the best results in bold and the second best results underlined. }\label{flash}
	\vspace{-1em}
	\begin{center}
		\begin{tabular}{c|ccccccccccccc}
			\hline
			\hline
			$\sigma^2$=25& Minion& Towel &Elmo & Pendant& Book & Tampax &Typewriter &Pot&Plant&Flower&Aloe&Cactus& Average\\ \hline	    
			CBM3D \cite{dabov2008image}  &33.63 &37.25 &36.03 &39.39 &36.17 &35.94 &34.37 &33.91 &33.82 &35.62 &31.86 &31.38 &34.95 \\
			DnCNN \cite{zhang2017beyond} &\underline{34.13} &\underline{37.58} &\underline{36.65} &\underline{39.78} &\underline{35.61} &\underline{36.55} &\underline{34.87} &\underline{34.35} &\underline{34.42} &\underline{36.26} &\underline{32.70} &\underline{31.62} &\underline{35.38}\\
			DJFR\cite{li2019joint}    &31.59&36.86&34.69&36.83&34.16&34.45&32.81&33.26&31.90&34.69&30.78&31.13&33.76\\
			MuGIF \cite{guo2018mutually} &30.49 &35.42 &33.75 &35.78 &32.62 &33.46 &31.51 &31.82 &30.95 &33.24 &29.49 &30.88 &32.45 \\
			CU-Net&\textbf{34.24} &\textbf{37.99} &\textbf{36.82} &\textbf{39.95} &\textbf{36.86} &\textbf{36.97} &\textbf{35.07} &\textbf{35.52} &\textbf{34.42} &\textbf{36.40} &\textbf{32.83} &\textbf{33.26}&\textbf{35.86} \\
			
			\hline					
			$\sigma^2$=50& Minion& Towel &Elmo & Pendant& Book & Tampax &Typewriter &Pot&Plant&Flower&Aloe&Cactus& Average\\ \hline		   
			CBM3D \cite{dabov2008image}  &29.94 &34.36 &32.50 &36.40 &\underline{33.15} &32.37 &31.30 &30.57 &30.17 &32.33 &27.98 &27.83 &31.58 \\
			DnCNN \cite{zhang2017beyond} &\underline{30.32} &\underline{34.65} &\underline{33.24} &\underline{36.58} &32.37 &\underline{32.95} &\underline{31.87} &\underline{30.87} &\underline{30.83} &\underline{33.04} &28.49 &\underline{28.04} &\underline{31.94}\\
			DJFR\cite{li2019joint}   &28.55&33.27&31.79&34.65&31.74&31.58&29.90&30.02&28.84&31.30&28.05&27.66&30.61\\
			MuGIF \cite{guo2018mutually} &26.93 &32.02 &30.94 &31.78 &29.72 &30.54 &28.57 &28.97 &27.45 &29.93 &\underline{28.97} &27.81 &29.22 \\
			CU-Net&\textbf{31.08} &\textbf{35.91} &\textbf{34.20} &\textbf{37.23} &\textbf{34.11} &\textbf{34.22} &\textbf{32.40}  &\textbf{32.88} &\textbf{31.16} &\textbf{33.60} &\textbf{29.17}&\textbf{30.69}&\textbf{33.05} \\
			\hline					
			$\sigma^2$=75& Minion& Towel &Elmo & Pendant& Book & Tampax &Typewriter &Pot&Plant&Flower&Aloe&Cactus& Average\\ \hline			
			CBM3D \cite{dabov2008image}  &27.85 &32.34 &30.75 &34.41 &31.40 &30.31 &29.56 &28.79 &28.20 &30.40 &25.83 &26.10 &29.66 \\			
			DnCNN \cite{zhang2017beyond} &\underline{28.33} &\underline{32.75} &\underline{31.16} &\underline{34.82} &\underline{31.83} &\underline{30.74} &\underline{29.97} &\underline{29.21} &\underline{28.46} &\underline{30.83} &\underline{26.25} &\underline{26.55} & \underline{30.08}\\
			DJFR\cite{li2019joint}   &26.65&31.77&30.63&32.78&29.54&29.86&28.85&28.33&27.54&29.25&25.56&26.26&28.92\\
			MuGIF \cite{guo2018mutually}&24.89 &30.30 &29.82 &29.70 &28.40 &28.54 &26.46 &27.44 &25.82 &27.84 &24.49 &26.27 &27.50 \\
			CU-Net                      &\textbf{29.12} &\textbf{34.25} &\textbf{32.55} &\textbf{35.12} &\textbf{32.93} &\textbf{32.19} &\textbf{30.55} &\textbf{31.66} &\textbf{29.17} &\textbf{31.73} &\textbf{27.03} &\textbf{29.30} &\textbf{31.30} \\
			
			\hline					
		\end{tabular}
	\end{center}
\end{table*}

\subsection{MIF related tasks}
\subsubsection{Multi-exposure image fusion}

\begin{figure*}
	\centering
	\centerline{\epsfig{figure=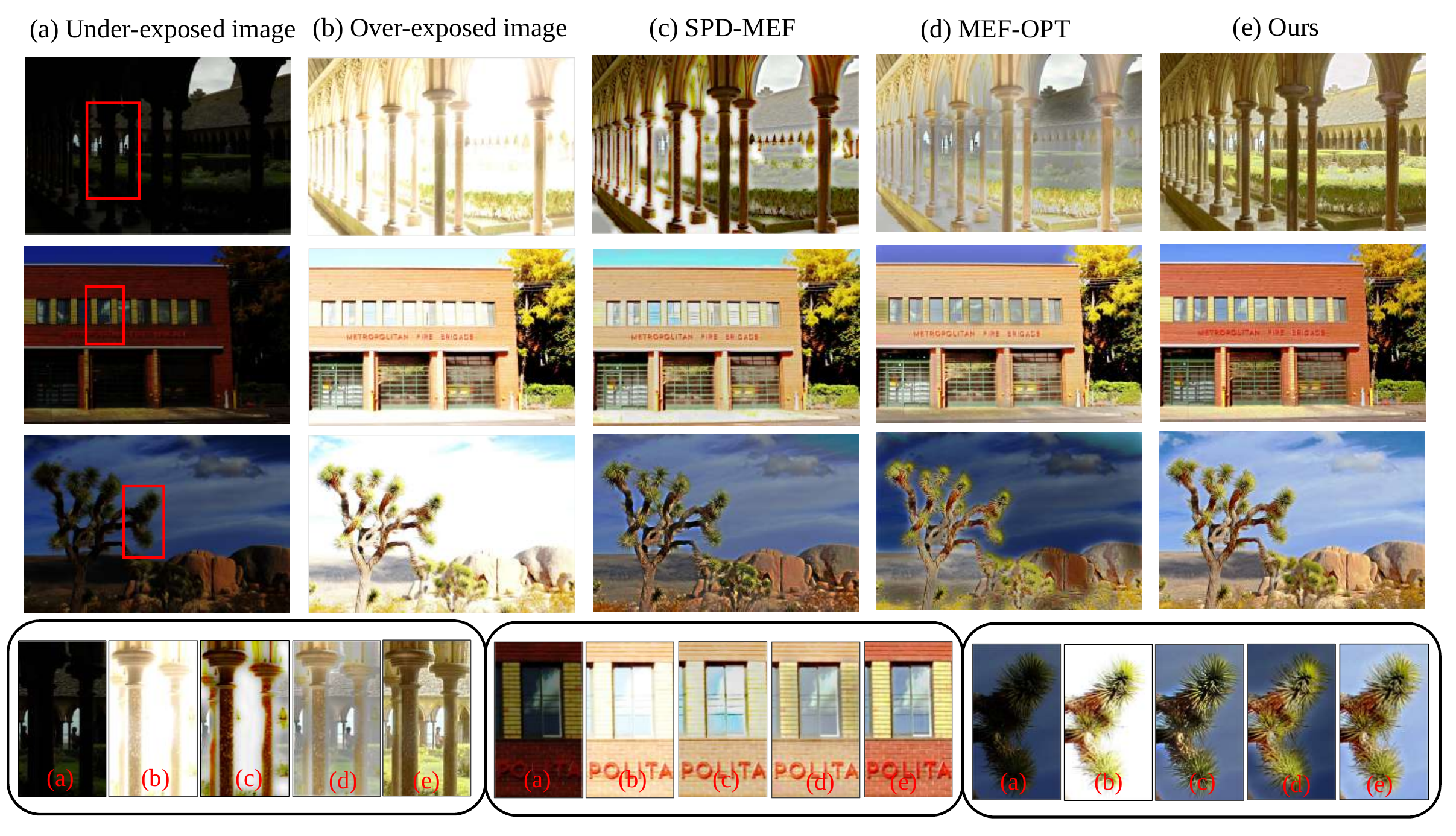, width=17cm}}
	\caption{Visual comparisons of multi-exposure image fusion results: (a)  input under-exposed image, (b) input over-exposed image, (c) SPD-MEF \cite{ma2017robust}, (d) MEF-OPT \cite{ma2018multi}, (e) our CU-Net.  The last row shows the enlarged regions in the corresponding images. Better view in electronic version. }\label{expo}
\end{figure*}

For the multi-exposure image fusion task, we aim to combine one under-exposed image with an over-exposed one to obtain a photo-realistic natural image.   We use the datasets from \cite{cai2018learning} for training and testing. The paper \cite{cai2018learning} provides the images with seven exposure levels, and we choose the first and sixth levels as the under-exposed and over-exposed images, respectively. Some examples of the images are shown in Fig. \ref{expo}. As we can see from this figure, the under-exposed image is extremely dark, and the over-exposed image is extremely bright, so that many details are lost in both of them. Our method is able to detect the useful information in these two images and re-assemble them to a new image which contains all the relevant information and looks visually pleasing. We compare our method with the other two state-of-the-art approaches on multi-exposure image fusion task: the SPD-MEF \cite{ma2017robust} and MEF-OPT\cite{ma2018multi}.  As we can see in Fig. \ref{expo}, the  SPD-MEF method suffers severe non-uniform brightness artifacts and contrast loss, and thus it loses many important details. For example, in the third picture of the first row, there should be a person standing between the two pillars, but the  SPD-MEF method fails to discover it. The MEF-OPT method performs better than SPD-MEF, but the brightness is still not uniform across the whole images. In addition, the combined image has  halo effects  around the edges (the fourth figure in the third row). In contrast, the images generated by our method are with consistent and uniform brightness across the whole image, and do not have  contrast loss or halo effects. 

\subsubsection{Multi-focus image fusion}
\begin{figure*}
	\centering
	\centerline{\epsfig{figure=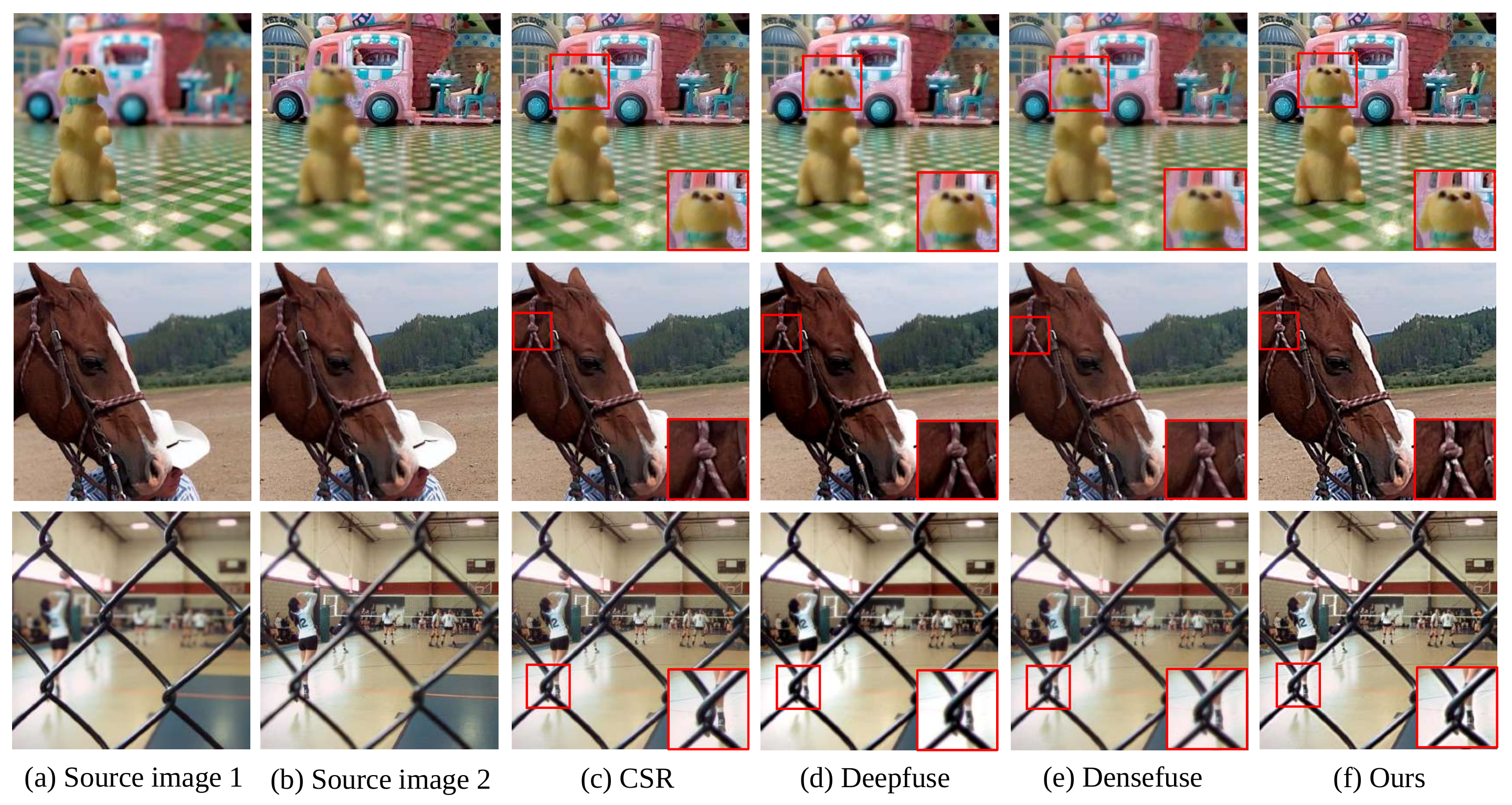, width=17cm}}
	\caption{Visual comparisons of multi-focus image fusion results. (a) The near-focus image, (b) The far-focus image, (c) CSR \cite{liu2016image}, (d) Deepfuse network \cite{prabhakar2017deepfuse}, (e) Densefuse network \cite{li2018densefuse}, (f) our CU-Net. Better view in electronic version.}\label{focuse}
\end{figure*}

Due to the finite depth-of-field of a camera, it is difficult to capture an image with all the objects in focus. The multi-focus image fusion task aims to fuse two or more  images focused on different depth planes to obtain an all-in-focus image. In this paper, we fuse one near-focus image with one far-focus image for an all-in-focus image. Examples of those images are shown in Fig. \ref{focuse}. The training dataset is from the General 100 dataset \cite{dong2016accelerating}. The near-focus and far-focus image pairs are generated by  blurring the randomly chosen foreground and background of an image using Gaussian blurring. The testing images are from the Lytro multi-focus image dataset \cite{nejati2015multi}. We compare our method with the following three state-of-the-art approaches:  CSR \cite{liu2016image}, Deepfuse network \cite{prabhakar2017deepfuse} and Densefuse network \cite{li2018densefuse}. The visual comparison results are shown in  Fig. \ref{focuse}. As we can see from this figure, the combined images generated by our method are more visually pleasant than the other approaches,  with every part in focus, and  with clear and sharp boundary edges. In contrast, the comparison methods, CSR \cite{liu2016image}, Deepfuse network \cite{prabhakar2017deepfuse} and Densefuse network \cite{li2018densefuse}, all lead to different levels of blurring artefacts across the boundary areas, as shown in the close-ups of the toy  $\textit{dog}$ and the $\textit{fence}$.

\subsubsection{Medical image fusion}
\begin{figure*}
	\centering
	\centerline{\epsfig{figure=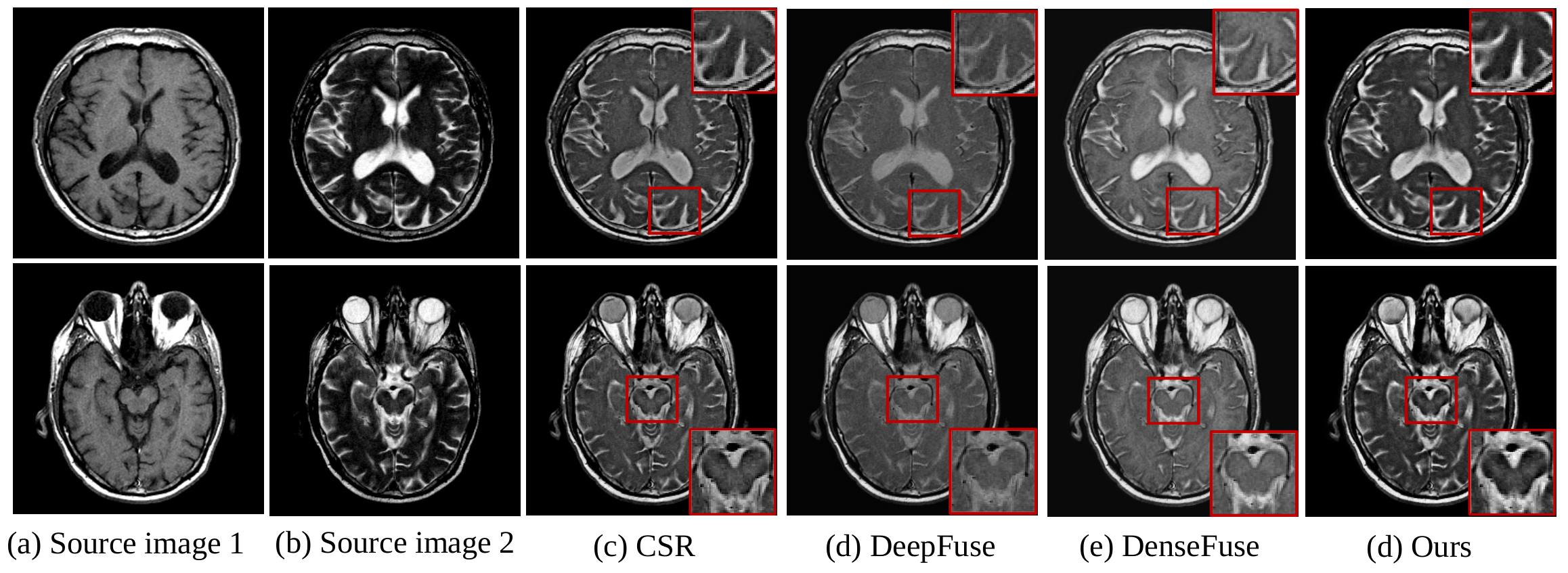, width=17cm}}
	\caption{Visual comparisons of medical image fusion results. (a) T1-weighted MR image, (b) T2-weighted MR image, (c) CSR \cite{liu2016image}, (d) Deepfuse network \cite{prabhakar2017deepfuse}, (e) Densefuse network \cite{li2018densefuse}, (f) our CU-Net. Better view in electronic version. }\label{med}
\end{figure*}

Medical image fusion plays an important role in clinical disease diagnosis. Since different medical images usually capture different features of the organs or tissues, a good combined image can deliver more useful information, which helps the effective diagnosis of some diseases. In this paper, we choose to fuse two different magnetic resonance (MR) images, i.e., the T1 weighted MR image and T2 weighted MR image. The testing dataset is from the Whole Brain Atlas \cite{summers2003harvard} from Harvard medical school. For deep learning based  medical image processing, one main problem is the lack of large training dataset. Here, we use the model trained by the multi-focus image fusion task and then  fine tune it using a small medical dataset which is composed of 30 image pairs from the  Whole Brain Atlas \cite{summers2003harvard}. Fig. \ref{med} shows the fused images using different methods, including CSR \cite{liu2016image}, Deepfuse network \cite{prabhakar2017deepfuse}, Densefuse network \cite{li2018densefuse} and our CU-Net. As we can see, the method CSR \cite{liu2016image} and Deepfuse network \cite{prabhakar2017deepfuse} both lose the image energy and have low contrast, which leads to low brightness in the soft-tissue regions. The method Densefuse \cite{li2018densefuse} produces over-bright images, which makes many details difficult to see. Our method is able to preserve well the structures and details of the organs and tissues without losing the image energy or producing the brightness distortions.

\begin{figure*}
	\centering
	\centerline{\epsfig{figure=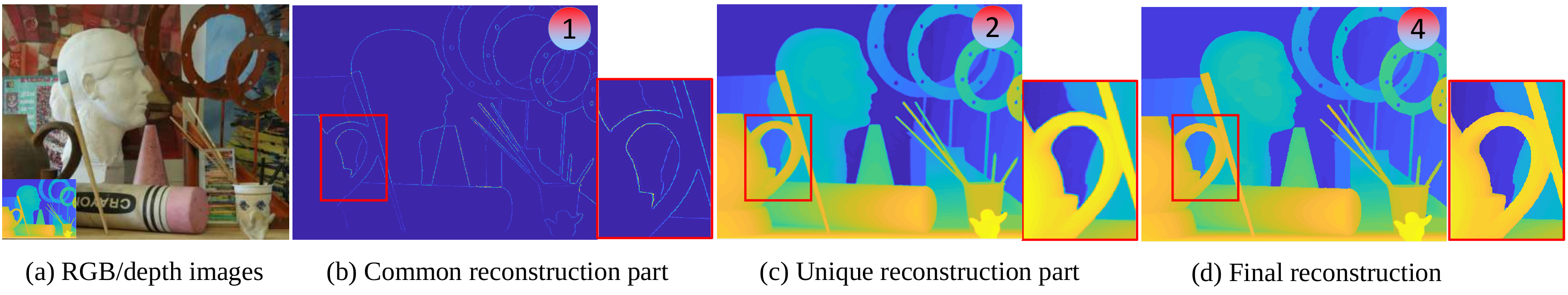, width=17cm}}
	\caption{Visualization of the reconstructed image of different parts in the CU-Net for the RGB guided depth SR task. (b) is the common reconstruction which  corresponds to Point 1 in Fig. \ref{frame}, (c) is the unique reconstruction, which  corresponds to Point 2 in Fig. \ref{frame}, and (d) is the final reconstruction, which is the sum of (b) and (c), corresponding to the Point 4 in Fig. \ref{frame}. Better view in electronic version.}\label{vd}
	\vspace{-.5em}
\end{figure*}

\subsection{Common and Unique Reconstruction Visualization}
As shown in Fig. \ref{frame}, the final reconstructed image is composed of one common reconstruction and one or two unique reconstructions (one for MIR related tasks and two for MIF related tasks).  To better verify the effectiveness of our method, we will visualize the common and unique components which contribute to the final reconstruction, i.e., the Point 1, 2, 3, and 4 in Fig. \ref{frame}.  For the MIR tasks, we take the RGB guided depth SR as example. Fig. \ref{vd} shows the common and unique reconstructed images for the RGB guided depth SR task. As we can see, the common reconstruction only contains the high-resolution edge information, which is consistent with the fact that the only common element between the RGB and depth images is the edge details. The depth information is entirely provided by the unique reconstruction, which is reasonable because the RGB image does not contain any depth information. However, in the unique reconstruction, depth discontinuities are still blurred. In the final reconstructed image, the blurring effect is successfully eliminated due to the addition of the common reconstruction which has sharp high-resolution edges. 

\begin{figure*}
	\centering
	\centerline{\epsfig{figure=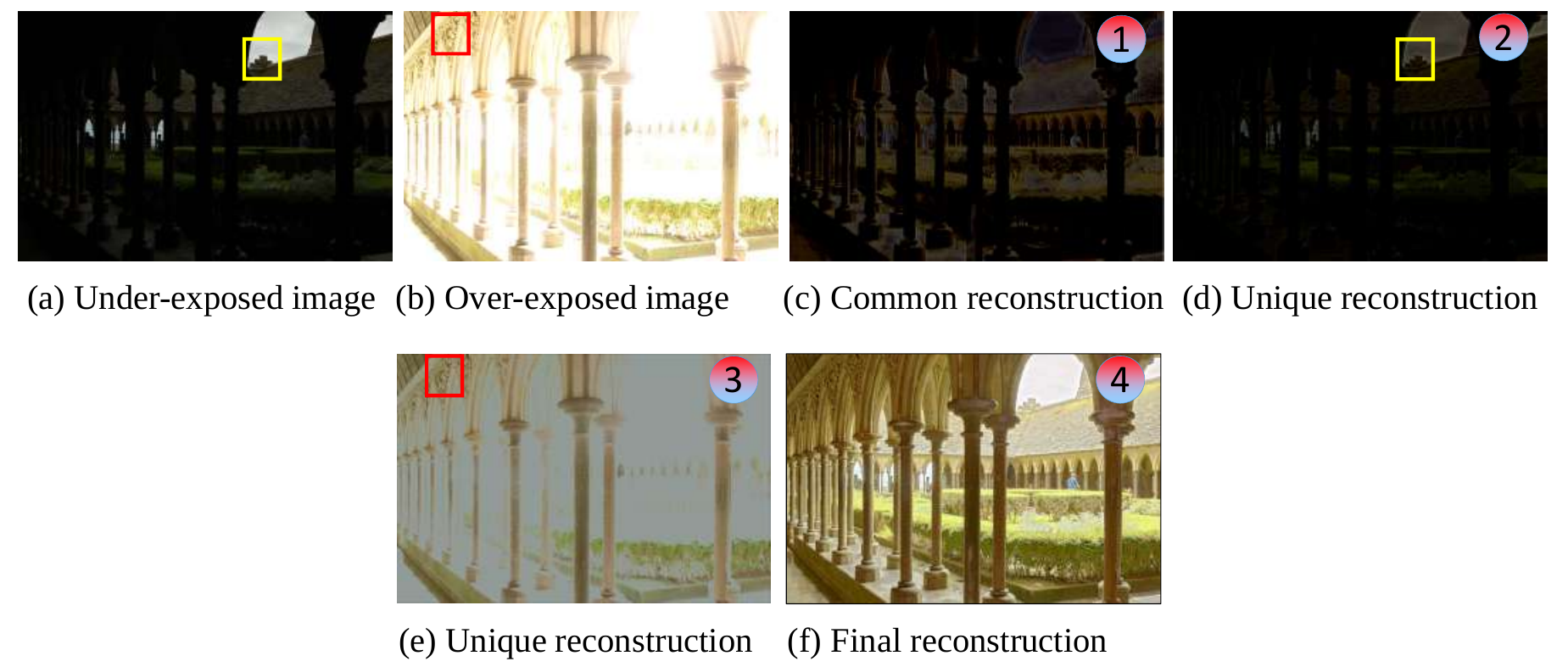, width=17cm}}
	\vspace{-1em}
	\caption{Visualization of the reconstructed image of different parts in the CU-Net for the multi-exposure image fusion task. (c) is the common reconstruction which corresponds to Point 1 in Fig. \ref{frame}, (d) and (e) are the unique reconstructions, which correspond to Point 2 and 3 in Fig. \ref{frame}, respectively,  and (f) is the final reconstruction, which is the sum of (c), (d) and (e). Better view in electronic version.}\label{ve}
		\vspace{-.5em}
\end{figure*}

For the MIF related tasks, Fig. \ref{ve} visualizes the three components of the final fused image for the multi-exposure fusion task. We can clearly see from this figure that the common reconstruction contains the common parts shared between the under-exposed and over-exposed images, while the two unique reconstructions preserve the unique features in each image, e.g., see the yellow and red marks in the images. By adding these three components together, we can obtain the final fused image, which contains both the common feature between the two extremely exposed images and the unique feature of each single image. 

%
\vspace{-0.5em}
\subsection{Ablation Study} \label{abh}
\textbf{Residual architecture.} As shown in Fig. \ref{module}, the UFEM and CFPM modules have residual structures. In the first ablation study, we explore the importance of this residual architecture on the reconstruction performance. To do this, we firstly remove the two residual arrows in the modules, which makes UFEM and CFPM pure feed-forward CNN modules. We then re-train the network using the same training strategy. Table \ref{red} shows the results on the RGB guided depth image SR task with and without the residual structures, respectively. As we can see, the residual architecture indeed helps increase the reconstruction accuracy. 

\textbf{Filter size.} We further analyse the effects of filter size $s\times s$ on the reconstruction performance. Table \ref{fs} shows the results on the RGB guided depth SR when we gradually increase the filter size $s$ from 3 to 9. We can see that the reconstruction performance is improved with larger  $s$. However, when the filter is too large, e.g., $s=9$, the reconstruction accuracy  decreases. The possible reason for this phenomenon is that a very large filter support may overlook local details, and this leads to less accurate reconstructions. In this paper, we choose to use a filter size of $8\times 8$.

\textbf{Network depth.}  The network depth also plays an important role in improving the reconstruction performance. Here, we use the number of LCSC blocks to indicate the network depth.  Table \ref{nd} presents the results with different numbers of LCSC blocks.  We can see that the reconstruction accuracy improves with the depth of the network, however,  the depth of the network  also increases the model size and training complexity. In this paper, we  use 4 LCSC blocks in each module, which represents a good trade-off between  reconstruction accuracy and training complexity. 

\begin{table}
	\caption{Importance of the residual module architecture.}\label{red}
	\vspace{-2em}
	\begin{center}
		\begin{tabular}{c|ccc} \hline \hline
			Architecture&  RMSE& PSNR (dB)& SSIM\\ \hline
			Residual&1.88&44.65&0.9936\\
			Non-residual&2.03&43.22&0.9898\\	 \hline				
		\end{tabular}
	\vspace{-1em}
	\end{center}
\vspace{-1em}
\end{table}

\begin{table}  \addtolength{\tabcolsep}{-3pt}
	\caption{Effects of the filter size.}\label{fs}
	\vspace{-2em}
	\begin{center}
		\begin{tabular}{c|ccccccc} \hline \hline
				Filter size& $s=3$ & $s=4$& $s=5$& $s=6$& $s=7$& $s=8$& $s=9$\\ \hline
			RMSE &2.13&2.05&1.96&1.93&1.90&1.88&1.92\\		 
			PSNR/ dB &43.45&43.94&44.24&44.39&44.51&44.65&44.44\\
			SSIM &0.9915&0.9921&0.9931&0.9934&0.9935&0.9936&0.9934\\ \hline
		\end{tabular}
	\end{center}
	\vspace{-2em}
\end{table}
\begin{table}  \addtolength{\tabcolsep}{-2pt}
		\vspace{-1em}
	\caption{Effects of the network depth in each module.}\label{nd}
	\vspace{-2em}
	\begin{center}
		\begin{tabular}{c|cccccc} \hline \hline
		Network depth& $1$ & $2$& $3$& $4$& $5$& $6$\\ \hline
		RMSE &2.07&1.98&1.94&1.88&1.84&1.81\\
		PSNR/ dB &43.88&44.18&44.35&44.65&44.90&45.12\\
		SSIM &0.9920&0.9930&0.9933&0.9936&0.9938&0.9940\\ \hline
		Model size /KB &66&99&132&165&198&231 \\ \hline		
		\end{tabular}
	\end{center}
	\vspace{-1em}
\end{table}
\begin{table*}  
	\caption{The average running time (in seconds) of our method on various tasks.}\label{rt}
	\vspace{-2em}
	\begin{center}
		\begin{tabular}{c|ccc|ccc} \hline \hline
			Task& Task 1 &  Task 2&  Task 3&  Task 4&  Task 5&  Task 6\\ \hline
			Image resolution &$1320\times1080$&$512\times512$&$900\times600$&$900\times600$&$520\times520$&$256\times256$\\
			Running time &0.85s&0.24s&0.38s&0.42s&0.27s&0.09s\\ \hline
			
		\end{tabular}
	\end{center}
	\vspace{-1em}
\end{table*}
	\vspace{-.5em}
\subsection{Running Speed}
For the real-time applications, the computational complexity of a method is an important factor to be considered. Our method is a deep learning based method using a  feed-forward network architecture, which has the advantage of fast running speed. In Table \ref{rt}, we present the running time of our method for the six applications presented in this paper. The task 1 to task 6 indicate the RGB guided depth image SR, RGB guided MS image SR, flash guided non-flash image denoising,  multi-exposure image fusion, multi-focus image fusion, and medical image fusion, respectively.  The time is recorded by implementing the experiments on a PC with an GEFORCE GTX 1080 Ti GPU. 
\vspace{-.5em}
\section{Conclusion} \label{con}
In this paper, we address the general multi-modal image restoration and image fusion problems by proposing  a novel and flexible CNN architecture, named Common and Unique information splitting network (CU-Net). Different from other approaches, our network architecture is derived from a new proposed multi-modal convolutional sparse coding (MCSC) model, which makes each part of our network interpretable. To verify the effectiveness of our CU-Net, we conduct exhaustive experiments on six different  multi-modal image restoration and  fusion tasks.  We also make a comprehensive ablation study  to explore the contribution of each component in the network and the effects of some critical network hyper-parameters, such as filter size and network depth. The experimental results show that our method outperforms other state-of-the-art methods in all the tasks considered,  with a small model size and fast running speed. 
\appendices 
\section{Learned Convolutional Sparse Coding \cite{sreter2018learned}} 

The typical convolutional sparse coding problem with $\ell_1$ constraint is formulated as follows:
\begin{equation} \label{app_ee1}
\underset{\{ \bm{u}_k \}}{\mathrm{Argmin}} \frac{1}{2}\left\|\bm{x}-\sum_k \bm{d}_k*\bm{u}_k \right\|_2^2
+\lambda \sum_k\left\|\bm{u}_k\right\|_1,
\end{equation}
where $\bm{x} \in \mathds{R}^{n\times n}$ is the input image, $\{\bm{d}^k\}_{k=1}^K  \in \mathds{R}^{s\times s}$ is a set of known filters, and $\{\bm{u}^k\}_{k=1}^K  \in \mathds{R}^{n\times n}$ is the filter response to be computed. Since the convolutional operation is linear, we can construct a Toeplitz matrix $\bm{D}_k \in \mathds{R}^{n^2\times n^2} $ to make $\bm{d}_k*\bm{u}_k$ $= \bm{D}_k \bm{u}_k'$ where $\bm{u}_k' \in \mathds{R}^{n^2}$ is the  vectorized $\bm{u}_k$. Then, we can turn \eqref{app_ee1} to the following:
\begin{equation} \label{app_ee2}
\underset{\{ \bm{u}_k' \}}{\mathrm{Argmin}} \frac{1}{2}\left\|\bm{x}'-\sum_k  \bm{D}_k \bm{u}_k' \right\|_2^2
+\lambda \sum_k\left\|\bm{u}_k'\right\|_1,
\end{equation}
where $\bm{x}'\in \mathds{R}^{n^2}$ is the vectorized  $\bm{x}$.
By concatenating all the $\bm{D}_k$, we can have $\bm{D} \in \mathds{R}^{n^2\times Kn^2}=\{\bm{D_1} \bm{D_2} \cdots \bm{D_K}\}$. By putting all the $\bm{u}_k'$ in a column, we can have  $\bm{u} \in \mathds{R}^{Kn^2}=\{\bm{u}_1'; \bm{u}_2'; \cdots \bm{u}_K'\}$. Then, Eq. \eqref{app_ee2} can be turned to a traditional sparse coding problem:
\begin{equation} \label{app_ee3}
\underset{ \bm{u} }{\mathrm{Argmin}} \frac{1}{2}\left\|\bm{x}'- \bm{D} \bm{u} \right\|_2^2
+\lambda \left\|\bm{u}\right\|_1.
\end{equation}
The Eq. \eqref{app_ee3} can be solved using the iterated shrinkage and thresholding algorithm (ISTA) \cite{daubechies2004}, which gives us the following iterative solution:
 \begin{equation} \label{app_ee4}
  \bm{u}_{j+1} =S_{\lambda}(\bm{u}_{j}+\bm{D}^T(\bm{x}'-\bm{D}\bm{u}_{j})).
 \end{equation}
 By replacing the matrix multiplication in Eq. \eqref{app_ee4} with convolutional operations, we can have the following:
  \begin{equation} \label{app_ee5}
 \bm{U}_{j+1} =S_{\lambda}(\bm{U}_{j}-\bm{E}*\bm{F}*\bm{U}_{j}+\bm{E}*\bm{x}),
 \end{equation}
 where $\bm{E}\in \mathds{R}^{s\times s \times K} $ and $\bm{F}\in \mathds{R}^{s\times s \times K} $ are the two sets of filters which lead to the Toeplitz matrices $\bm{D^T}$ and $\bm{D}$, respectively, and $\bm{U} \in \mathds{R}^{n\times n \times K}$ is the stack of  $\{\bm{u}^k\}_{k=1}^K$ to be learned in Eq. \eqref{app_ee1}. In the learned convolutional sparse coding (LCSC), the   $\bm{E}$ and  $\bm{F}$ are learnable as convolutional layers  in a deep network.
\bibliographystyle{IEEEtran}
\bibliography{IEEEfull,nips18}

\end{document}